\documentclass[lettersize,journal]{IEEEtran}
\usepackage{amsmath,amsfonts}
\usepackage{algorithmic}
\usepackage{algorithm}
\usepackage{array}
\usepackage[caption=false,font=normalsize,labelfont=sf,textfont=sf]{subfig}
\usepackage{textcomp}
\usepackage{stfloats}
\usepackage{url}
\usepackage{verbatim}
\usepackage{graphicx}
\usepackage{tabularx}
\usepackage{array}    
\usepackage{subcaption}
\usepackage{cite}
\usepackage{caption}
\usepackage{subcaption}
\usepackage{booktabs}
\usepackage{xcolor}
\usepackage{placeins}

\begin{document}




\title{Experimental investigation of pose informed reinforcement learning for skid-steered visual navigation.}

\author{Ameya Salvi,~\IEEEmembership{Student Member,~IEEE,}~Venkat Krovi,~\IEEEmembership{Senior Member,~IEEE} 
\thanks{DISTRIBUTION STATEMENT A. Approved for public release; distribution is unlimited. OPSEC\#9334},%
\thanks{All authors are with the Department of Automotive
Engineering at the Clemson University International Center
for Automotive Research (CU-ICAR), Greenville, SC 20607.\{asalvi,vkrovi\}@clemson.edu}
}



\maketitle

\begin{abstract}

Vision-based lane keeping is a topic of significant interest in the robotics and autonomous ground vehicles communities in various on-road and off-road applications. The skid-steered vehicle architecture has served as a useful vehicle platform for human controlled operations. However, systematic modeling, especially of the skid-slip wheel terrain interactions (primarily in off-road settings) has created bottlenecks for automation deployment. End-to-end learning based methods such as imitation learning and deep reinforcement learning, have gained prominence as a viable deployment option to counter the lack of accurate analytical models. However, the systematic formulation and subsequent verification/validation in dynamic operation regimes (particularly for skid-steered vehicles) remains a work in progress. To this end, a novel approach for structured formulation for learning visual navigation is proposed and investigated in this work. Extensive software simulations, hardware evaluations and ablation studies now highlight the significantly improved performance of the proposed approach against contemporary literature. Code : https://github.com/ameyarsalvi/PoseEnhancedSSVN.git

\end{abstract}


\section{Introduction}~\label{sec:introduction}

Lane-keeping or lane-following is a key autonomy capability for successful wheel mobile robot (WMR) deployments for on-road (and increasingly in manufacturing shop floor and off-road) applications. Vision enhanced navigation has long been a subject of interest since the earliest autonomous mobile robot, Shakey (circa 1970)~\cite{Nilsson1984ShakeyTR}. The subsequent growth in popularity has benefited from both the scientific progress (new deterministic and probabilistic methods), and the technological progress (convergence of computing, communication and miniaturization). 
In as much, it has proved to be a fertile ground for cross-disciplinary research in real-time perception, planning, and control within increasingly size-, weight-, and power- (SWaP) limited computation and communication frameworks. 

\begin{figure}
    \centering
    \includegraphics[width=1\linewidth]{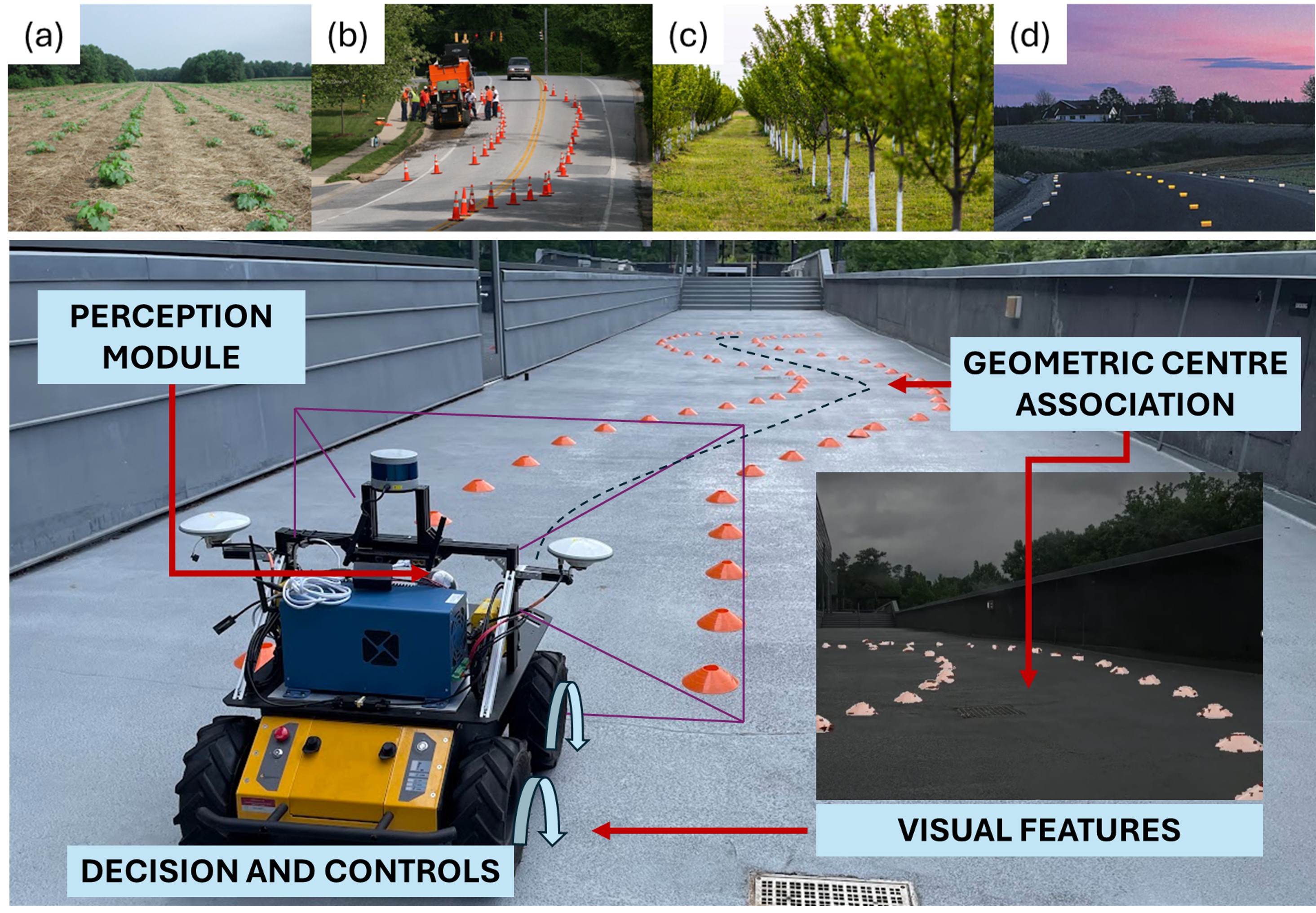}
    \caption{ (Top Row) Several scenarios where lanes are either created or realized by sparse visual markers : (a) Rows of crops, (b) Cones for lane modifications, (c) Trees in vineyards, and, (d) radium road markers .
    (Bottom Row) A typical vision based path following framework involves (i) capturing and processing vision data which allows the extraction of meaningful environmental features, (ii) associating those features in context to the relevant robot control task, and, (iii) planning and execution of the control task based on the formerly determined association. }
    \label{fig:overview}
\end{figure}

Over the years, this research has matured from deployments in low-speed indoor applications with small-scale robots to increasingly complex, unstructured and uncertain environments~\cite{DeSouza2002,Bonin-Font2008}. Some dimensions of the growth of complexity include: robust perceptual processing (identifying markers, lanes, and patterns), size (small-, mid-, and full-scaled vehicles), vehicle speed (limits of tire dynamics), and computational processing (data transfer bottlenecks). Beginning in the 1980s, visual servoing emerged as a robust and resilient approach for coupling perception to planning to action. Fig.~\ref{fig:overview} illustrates one such exemplary lane following problem posed as a outdoor visual navigation task involving aspects of perception, planning, and control.

As noted in the Section~\ref{sec:literature}, the underlying nature of the vehicle platform is coupled with the complexity of the operational environment to determine the performance of visual servoing as shown in~\ref{fig:overview}. For example, deployments in outdoor road settings face perceptual processing challenges (weather, time of day, texture, terrain and lane boundary structure). Similarly, the choice of underlying vehicle configuration (Ackermann or differential drives, various suspension subsystems choices) also impact the performance of the visual servoing task. In particular, a unique challenge is faced by skid-steered wheel mobile robots (SSWMRs) in outdoor unstructured settings.



SSWMRs have emerged as a very useful architecture for deployments in indoor, and increasingly outdoor low-speed applications owing to their simplicity of controls, lack of suspensions, ability to execute zero radius turns, among many other attributes, but have required expert human skills to extract performance. Hence, SSWRMs have largely required human participation for both in-situ operation, as well as remote teleoperation. Capturing the vehicle's performance in suitable predictive motion model is crucial for achieving partial or fully autonomous operations. 
However, SSWMRs posses unique motion mechanics which depend on wheel skidding, i.e. explicit violation of both the longitudinal (pure roll) constraints, as well as the lateral (no side slip) constraints that characterizes the traditional disc wheels. The uncertain/ unexpected constraint violations now compromise adequate predictive models and thereby inclusion in subsequent model-based control frameworks.  


Deep learning (DL), and in particular, deep reinforcement learning (DRL) has provided a viable solution for robot control tasks lacking explicit predictive models by leveraging high fidelity simulators. Learning based methods bypass the temporal constraints posed by on-line optimization approaches by facilitating \textit{off-line} learning for an optimal policy that can be deployed \textit{real-time}. Thus, the contribution of this effort is from the pursuit of deploying learning based approaches for visually servoed SSWMRs including:


\begin{itemize}
    \item \textbf{Formulation}: Formulating a pose informed DRL framework for lane keeping and its evaluation via parametric studies for performance improvement in simulation (investigating the distance between waypoints) and zero-shot policy deployment in reality (waypoint selection and sensor dropout).
    \item\textbf{Evaluation}: Evaluating the framework in simulation and hardware deployment with classical and contemporary lane centering approaches such as model predictive control and Pure pursuit.
    \item \textbf{Generalization}: Investigating the performance of the framework for utility beyond training markers (cones) to real lane following scenarios.
\end{itemize}

The rest of the paper has been organized as follows - Section~\ref{sec:literature} captures the relevant literature in the areas of vision based path following in the context of structured and unstructured environments, discusses recent trends in the literature and presents the existing research gaps. Section~\ref{sec:problem_formulation} describes problem formulation and the research methodology. Sections~\ref{sec:experiments-sim} and~\ref{sec:experiments-deploy} discuss various aspects of the experimental protocol development before presenting the results (both in simulation and deployment). Section~\ref{sec:generalization} tries to identify utility of the trained policy for zero-shot transfer to unseen scenarios in absence of cones (such as complete and partial lane boundaries). Finally, Section~\ref{sec:discussion} summarizes the overall work and presents potential avenues for further investigation.

\section{Related literature}~\label{sec:literature}

The literature surrounding the area has been organized to capture the evolution of visual navigation for mobile robots over the past fifty years. The three subsections in this section cover: (i) classical visual servoing (vision based control); (ii) its progression into indoor and outdoor visual navigation (complementary modular planning and control); and (iii) learning based visual navigation (end-to-end planing and control).

\subsection{Classical visual servoing}~\label{subsec:classic_vs}
Utilizing vision as an essential perceptual modality for robot control was kick-started by early advances in image processing technologies such as edge detection (Canny/ Sobel) and Hough transform~\cite{canny1986computational,ballard1981generalizing}.
This gradually led to the formalization of the topic of robotic visual servo-control ~\cite{Hutchinson1996,corke2011robotics} and subsequent categorization into Pose-Based Visual Servoing (PBVS), and Image-Based Visual Servoing (IBVS). 
Early developments of visual servo control of WMRs relied upon feature extraction using existing image processing technologies but with a focus much more on analytical modeling and control of the feature dynamics (relation between motion of wheels to image features). Such an approach allowed for pixel level feature manipulation in the image frame achieved by precise control of the wheel motions~\cite{Coulaud2006Stability,Mariottini2007epipolar,Ma1999vision,Zhang2002Visual,Nierobisch2006PanTilt,Chen2006homography}.

Over the years, several seminal works have contributed to WMR visual servoing by developing analytical solutions to the visual servoing problem, analyzing the stability of controllers and considering control in the presence of different camera geometries (pan-tilt and omni-directional cameras)~\cite{Nierobisch2006PanTilt,vidal2003formation}. In recent times, the challenges of robustness associated with field of view constraints and loss of the tracking objects~\cite{fu2013visual} have also gained prominence. Inasmuch, visual servo approaches catered much more towards the feature tracking problem rather than addressing autonomous navigation and its engendered nuances.


\subsection{Indoor and outdoor visual navigation}~\label{subsec:in-out-visnav}

The holistic autonomous navigation challenge was relieved to an extent by integration of the visual servo controllers within a robot navigation stack which functioned along side the localization and planning modules~\cite{DeSouza2002,Bonin-Font2008}. These methods capture some way of estimating the location of the robot in the environment to guide the robot towards desired goal (map-based visual navigation methods)~\cite{Kosaka1992FastVM,Meng1993MobileRN,Kabuka1987PositionVO}, or, incorporated techniques such as optic flows for feature tracking and following (map-free visual navigation)~\cite{Santos1993}. Yet, all of these methods rely heavily on the accuracy of the feature dynamics and  are limited to controlled indoor environments. The typical motion models utilized in these works are non-holonomic kinematic robot models, which rarely incorporate challenges associated with inertial dynamics and terramechanics thus limiting the scope to low-speed indoor mobile robotics applications.

Parallely, outdoor visual navigation techniques achieved significant development for structured environments with projects such as NAVLAB~\cite{Thorpe1988}, VITS~\cite{truk1988vits}, ALVINN~\cite{jochem1995vision} and Promethus~\cite{graefe1992vision}. These methods primarily followed the idea of map-less visual navigation, where road features (specifically lanes) were extracted using methods such as the Hough transforms and then used for downstream control tasks. These methods were further perfected and incorporated within complete autonomous navigation stacks used in DARPA programs such as the PerceptOR~\cite{Kelly2006PerceptOR}, LAGR~\cite{Jackel2006LAGR} and Grand challenge~\cite{thrun2006stanley}. Many of the methods also made way in partial autonomy stacks in the form of advanced driver assistant systems (ADAS) such as lane keeping assistance or lane departure warning features in the modern day automobiles~\cite{Bian2020,Hsiao2009}. While outdoor visual navigation techniques were a significant upgrade over the indoor methods, they still faced limitations in challenging operation domains. For instance, the Hough transform method tries to extract straight lines as image features which work well for straight highway routes but starts facing limitations in roads with higher curvatures. Further, the model based controllers within the navigation stack require heuristically tuning methods such as gain scheduling or rule based parameter selections to incorporate various operating conditions such as varying speeds and curvature. While some of these lower level control challenges are being alleviated using model predictive control (MPC) strategies in the ADAS technologies~\cite{hu2020lane,Fang2024}, the challenges associated with relevant control feature extraction is still an active research area~\cite{Frey2024RoadRunner}. Further, the solution for an MPC task greatly depends on the compute capabilities of the robot, which may vary depending on the robot's form factor and its ability to carry edge computers. While it is valuable to note that MPC by itself has shown tremendous success for trajectory tracking tasks for different robot platforms, it faces significant challenges when coupled with the process of generating waypoints purely from vision data as shown in further sections of this work.

\subsection{Learning for visual navigation}~\label{subsec:learn-visnav}

Over the last decade, the challenges associated with feature extraction, non-linear controls and compute complexities have been remarkably diminished with the aid of powerful machine learning algorithms. More specifically, deep learning utilizing convolutional neural networks (CNNs) are trained to map image pixels (input road images) to control actions such as acceleration and steering (collected from an expert driver). The idea was successor of the project ALVIN~\cite{jochem1995vision} and was scaled it to more challenging scenarios with contemporary developments in the learning technologies NVIDIA~\cite{bojarski2016end}. This approach of training a control policy is know as imitation learning (IL) which relies on an expert driver to first collect the training data. Thus, the performance of the policy can be limited by the performance of the skilled data collector. Such a challenge of unbiased data collection/generation is alleviated by deep reinforcement learning (DRL).


DRL is an iterative learning strategy that learns an expert driving policy from large streams action-observations sequences in a self-supervised manner~\cite{sutton2018reinforcement}. The idea of utilizing DRL for visual navigation entails using camera images of the road as a preview and learning wheel and steering actuations for keeping the robot on the desired path. The preliminary implementation of the idea could be found as early as 2000s~\cite{Se-Young2000}, but much more advancements (with developments in neural network architectures and novel learning agents) could be found in recent years~\cite{Chen2019,Liang2018,Toromanoff2020,Jaritz2018}.

The iterative learning aspect of DRL requires large streams of data to learn a stable learning policy. Further, the initial stages of the learning process encourages the agent to undertake exploratory actions to canvas the action-observation state space, which can be hazardous to human life and property. These two factors primarily make it extremely difficult to train the learning policy on the physical robot and lead way to the rise of robot simulator packages to be utilized for DRL~\cite{Todorov2012,isaacsim,coppeliasim,Egli2024RLExcavation}. It can be observed that while~\cite{Se-Young2000} did show case physical deployment of the DRL policy (both training and deployment), the majority of works in recent times have been \textit{simulation} only~\cite{Chen2019,Liang2018,Toromanoff2020,Jaritz2018,salvi2022virtual}. While simulation packages do serve as a decent surrogate for the physical robot, they fail to capture some crucial aspects of robot dynamics which results in what the community terms as the Sim2Real (simulation to reality) gap. While there have been some scaled version implementations of the problem~\cite{stable-baselines3,Li_2023}, they have been constrained to small vehicles or controlled environments, where the sim2real gap is at the minimum. As the scale of the robot platforms increase, effects of friction and non-linear tire dynamics start to broaden this reality gap leading to sub-optimal performance of the trained policies when transferred from simulation to reality~\cite{Zhao2020a,Eser2022}. 

To summarize, the investigation of deep reinforcement learning based physical deployments for robots with large sim2real gap is still a significant research topic, insights driven from which can contribute towards both platform agnostic visual navigation policy learning (visual navigation for all-wheel or Ackermann steered platforms), and, platform specific control policy learning (sim2real navigation policy transfer for SSWMRs). To this end, the problem of learning a robust policy for structured outdoor visual navigation is addressed in this work. Clearpath Husky~\cite{clearpath_husky}, as a representative SSWMR has been chosen as the deployment platform representing the wide simulation to reality gap presented by the skidding requirements for the robot motion.

\begin{figure*}[t]
    \centering
    \includegraphics[width=\textwidth]{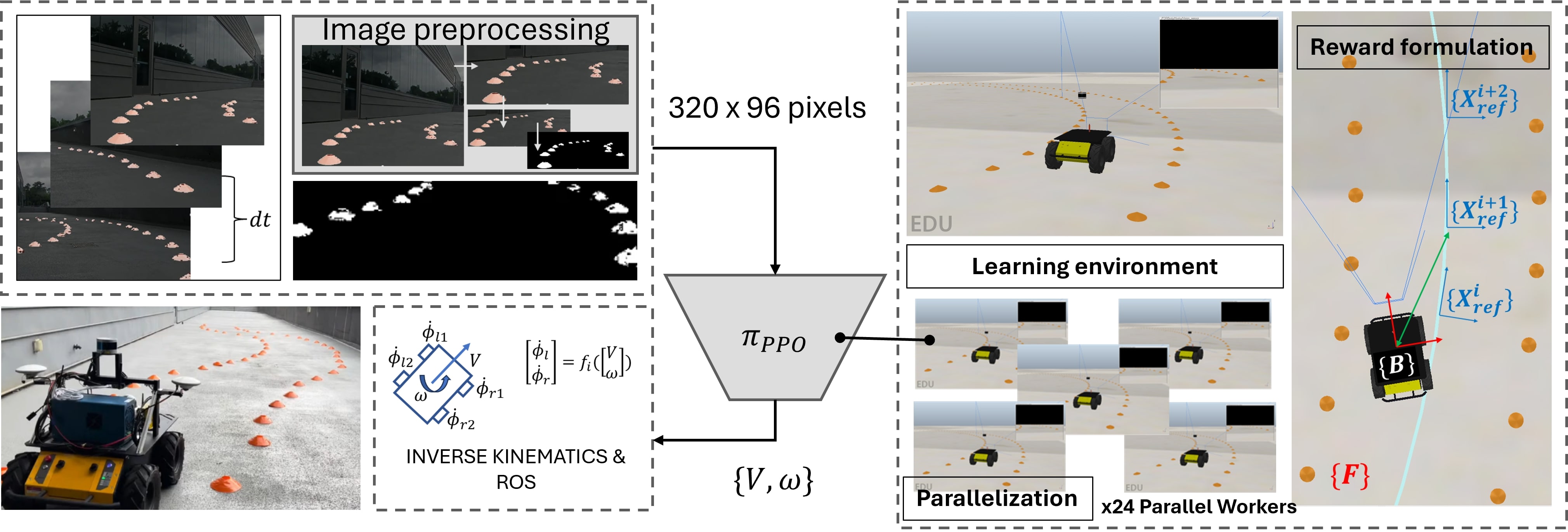}
    \caption{Schematic overview of the visual navigation problem in the form of driving withing lane markers.(A) Left : The deployment phase in which preview images are simplified to distill lane features, serving as input to the control policy that generates reference velocities in the robot body frame. The reference velocities are tracked using an inverse kinematics model, tuned from physical data. (B) Right : The back-end learning process which involves mapping the input images to geometric error between lane centre and robot centre (discussed in detail in section~\ref{subsec:guided_learning})}
    \label{fig:DRLOverview}
\end{figure*}

\section{Problem Overview}~\label{sec:problem_formulation}

This section covers the three key aspects of the proposed learning framework. First, the formulation section captures the essential notations and necessary abstractions in realizing the physical navigation task as a learning problem. Next, the proposed learning methodology is discussed highlighting the nuances explored in this investigation. Finally, the section concludes with establishing the crucial training hyper-parameters for policy learning.


\subsection{Formulation}



Figure~\ref{fig:DRLOverview} illustrates the realization of the visual navigation problem discussed in this work. Consider a SSWMR tasked with traversing a path which could be distinguished with some lane markings. $\{\mathcal{F}\} ~ \text{and}~\{\mathcal{B}\}$ define the spatial frame and robot body frame respectively. At any instance, the three dimensional robot pose $\mathbf{X} (:=[x,y,\theta])$, and the reference tracking pose $\mathbf{X}_{ref}$ are defined in spatial frame $\{\mathcal{F}\}$. ${}^{B}V, {}^{B}\omega~\text{and}~{}^{B}\dot{\phi}_{i}(i\in[1,4]) $ are the longitudinal velocity, yaw rate and wheel velocities of the robot, all expressed in robot body frame $\{\mathcal{B}\}$. Such a distinction in state representation significantly simplifies the problem formulation and evaluations as the robot pose is typically available in a fixed local or a global frame (GPS or odometery measurements) where as, the robot velocities are easier to be captured with filtered inertial measurement unit (IMU) readings, available typically in robot body frame.

The path is represented as orange soccer cones instead of well defined lane markings. This variant of the problem represents the fact that clear tractable lines may not be always available and path could be identified only as environmental patterns such as rows of crops, or routes, or guide ways in mines and construction sites. The intermittent breaks in the cone placements make it difficult for most of the standard lane identification tools making conventional approach towards lane identification significantly difficult. The input images are sampled at $dt$ seconds, and, minimally processed to distinguish the markers from the background before being sent to the control policy. The raw input image of the size $[640,480,3]$ is simplified to a binary image of the size $[320,96]$, where each pixel in the matrix can take a value between 0 and 255. At any time, $t$, the entire  feature matrix is represented :

\[ \mathbf{I_{t}} \in \mathbb{R}^{320 \times 96}\]

The feature matrix, $\mathbf{I_{t}}$, is further distilled using the Atari CNN network to distill the images to a size $64 \times1$ vector~\cite{mnih2015human}. The idea behind such distillation is to reducing the training complexity and encode only the relevant track information. The distilled features are part of the state matrix ($\mathbf{s_{t}}$) at any time $t$, which within the scope of this problem includes the robot positions in spatial frame ($[x, y, \theta]$), and, robot linear velocity in bodyframe, ${}^{B}V$, measured at the robot's center of gravity (CG), given as :

\[ \mathbf{s}_{t} = [x_{t},y_{t},\theta_{t},{}^{B}V_{t}, \mathbf{I}_{t}] \]


The action \( \mathbf{a_{t}} \) can be represented as a $2 \times 1$ vector:
\[
\mathbf{a}_{t} = \begin{bmatrix}
{}^{B}V_{a_{t}} \\
{}^{B}\omega_{a_{t}}
\end{bmatrix},~\forall~{}^{B}V_{a_{t}} \in [0.1, 1.0]~\&~{}^{B}\omega_{a_{t}} \in [-0.5, 0.5]
\]

Physically, ${}^{B}V_{a_{t}}~(m/s) $ and ${}^{B}\omega_{a_{t}}~(rad/s) $ represent the reference linear and the angular velocity set-points in the body frame. A reward function, $\mathbf{R}(\mathbf{s}_{t},\mathbf{a}_{t})$ can be defined as function of states and actions. The states, actions, reward along with the state transition probability, $\mathbf{p}$, and the discount factor, $\mathbf{\gamma}$, constitute the Markov Decision Process (MDP) tuple, given as : $\mathbf{M} = (\mathbf{s_{t}},\mathbf{a_{t}},\mathbf{p},\mathbf{R},\mathbf{\gamma})$. The objective of the learning agent is to learn an optimal policy, $\mathbf{\pi}^{*}(obs_{t})$, that generates action $\mathbf{a_{t}}$, given observations ${obs}_{t}$ (${obs}_{t} \in {\mathbf{I}_{t}}$) , subject to maximization of an expectation, $\mathbb{E}$, which is summation of the discounted reward:

\[
\pi^*(obs) = \arg\max_{\pi} \mathbb{E}_{\pi} \left[ \sum_{t=0}^{\infty} \gamma^t \mathbf{R}(\mathbf{s}_t, \mathbf{a}_t) \right]
\]

The state transition probability, $\mathbf{p}$, also known as the state transition dynamics, defines how the system's states transition from one time step to the next under the influence of action, $\mathbf{{}a_{t}}$. These state transitions are implicitly learnt by the learning agent through repeated interactions with the learning environment. The learning environment is an object that receives actions provided by the agent and returns observations to the agent as progressed states. In essence, an environment is the physical system or its digital replica for which a control policy is being trained for. In this problem, the environment is the combination of the skid-steered robot along with the physical environment in which the lane markers are placed. As discussed earlier, training directly on the physical system can be challenging, for which, CoppeliaSim robotics simulator is used as surrogate for the learning environment. In seminal works, such surrogate learning environments have also come to be know as gymnasiums (gym) because the agent \textit{trains} in those environment.

\subsubsection{Robot kinematics}~\label{subsubsec:IK}

While DRL methods excel in scenarios lacking analytical robot models, augmenting the learning process with partially known or approximate models can simplify it significantly. For SSWMRs, kinematic models, such as the extended differential drive model, are widely used for feedback control or as inverse kinematics (IK) controllers~\cite{Huskic2019Highspeed,Mandow2007,Yi2007a,Rabiee2019,Pazderski2017,Wang2023,trivedi2024,salvi2024online,salvi2025characterizing}. These models treat the skid-steered robot as a kinematic differential drive system, augmented with tunable factors learned from sensor data. Alternatively, some approaches fit linear or standard non-linear functions, like Gaussian models, to experimental data~\cite{Baril2020}.

Dynamic modeling studies for skid-steered robots~\cite{Yu2009,Caracciolo1999,Chuy2009,Kozowski2004,Aguilera-Marinovic2017,McCormick2022} often assume small robots in controlled environments, limiting their generalizability. Challenges also arise from relying on off-the-shelf robots, where low-level control maps (e.g., torque-voltage relationships) are unavailable, necessitating high-level control commands.

In this work, a mid-scale Clearpath Husky~\cite{clearpath_husky}, operating at speeds up to 1 m/s, is used. Its operational conditions are known a priori, enabling the use of an IK model fitted through standard approaches. A variant of the extended differential drive model is adopted as the nominal basis for fitting :


\begin{equation}
\label{eq:IKModel}
\begin{bmatrix}
    {}^{B}V \\
    {}^{B}\omega
\end{bmatrix}
= \frac{\hat{r}}{2}
\begin{bmatrix}
    1 & 1 \\
    -1/\hat{b} & -1/\hat{b}
\end{bmatrix}
*
\begin{bmatrix}
    {}^{B}\dot{\phi}_{L} \\
    {}^{B}\dot{\phi}_{R}
\end{bmatrix}
\end{equation}

Where, ${}^{B}V$ and ${}^{B}\omega$ are the measured longitudinal linear and the planar angular velocity of the robot in its body frame expressed as a function of the measured left and right wheel velocities, ${}^{B}\dot{\phi}_{L}$ and ${}^{B}\dot{\phi}_{R}$. For a typical differential drive robot, $b$ and $r$ constitute the track width and wheel radius, which over here are represented as $\hat{b}$ and $\hat{r}$ to indicated that they will be used as model fitting parameters to learn an \textit{estimated} track width and an \textit{estimated} wheel radius, based on experimental data. To limit the scope of this study, the robot is ensured to be near symmetric (mass distributions and same tire-terrain interactions for all wheels), relatively non-varying operation speeds and moving in know terrains, under which assumptions, lateral skidding at the robot's geometric center will be absent or is negligible. 


While it is intuitive that such a simplification should accelerate the model identification process and thus potentially improving the sim2real transfer, it is crucial to investigate the impact on the learning performance due to such a simplification. To this end, the learning performances between an end-to-end learning regime (image to wheel control) and inverse kinematics based learning regime (image to reference body velocities) is a subject of investigation in this work. Such a modification presents two variations of learning scenario, where the actions could be either reference body velocities or reference wheel velocities :

\begin{align}
\label{eq:action-IK-E2E}
    \mathbf{a}_t = 
    \begin{cases} 
      \begin{bmatrix}
      {}^{B}V_{a_{t}} \\
      {}^{B}\omega_{a_{t}}
      \end{bmatrix}, & \text{if using inverse kinematics} \\[10pt]
      \begin{bmatrix}
      {}^{B}\dot{\phi}_{L,a_{t}} \\
      {}^{B}\dot{\phi}_{R,a_{t}}
      \end{bmatrix}, & \text{if using end-to-end learning}
    \end{cases}
\end{align}

It is useful to note that a superior and robust performance for velocity tracking can be achieved by introducing a feedback controller. But, inclusion of such a controller entails introducing one more sensor (IMU or GPS) within the control framework for accurately capturing the robot velocity. Thus, to avoid additional hardware and compute overhead, a purely vision based control methodology is explored in this work.

~
\subsubsection{Learning track}~\label{subsec:track}
\begin{figure}
    \centering

    \begin{minipage}{0.95\linewidth}
        \centering
        \includegraphics[width=\linewidth]{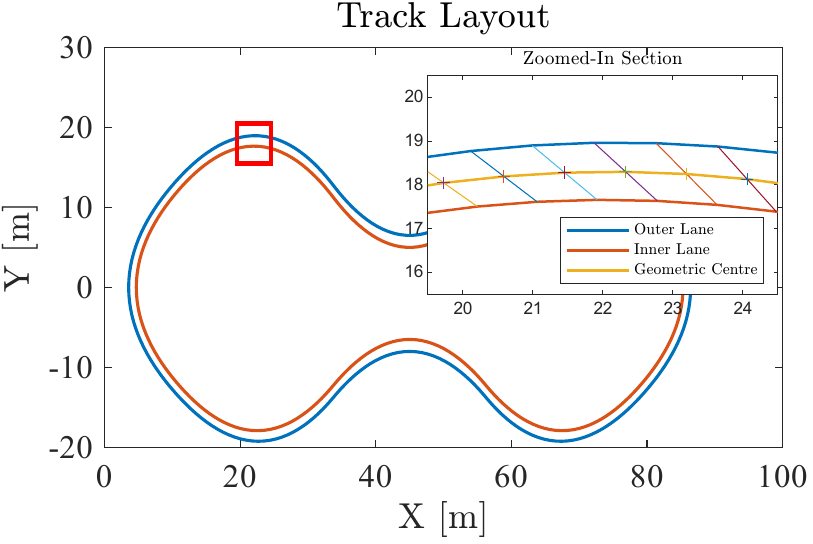}
        \vspace{0.3em}
        \small (a)
    \end{minipage}

    \vspace{0.7em}

    \begin{minipage}{0.95\linewidth}
        \centering
        \includegraphics[width=\linewidth]{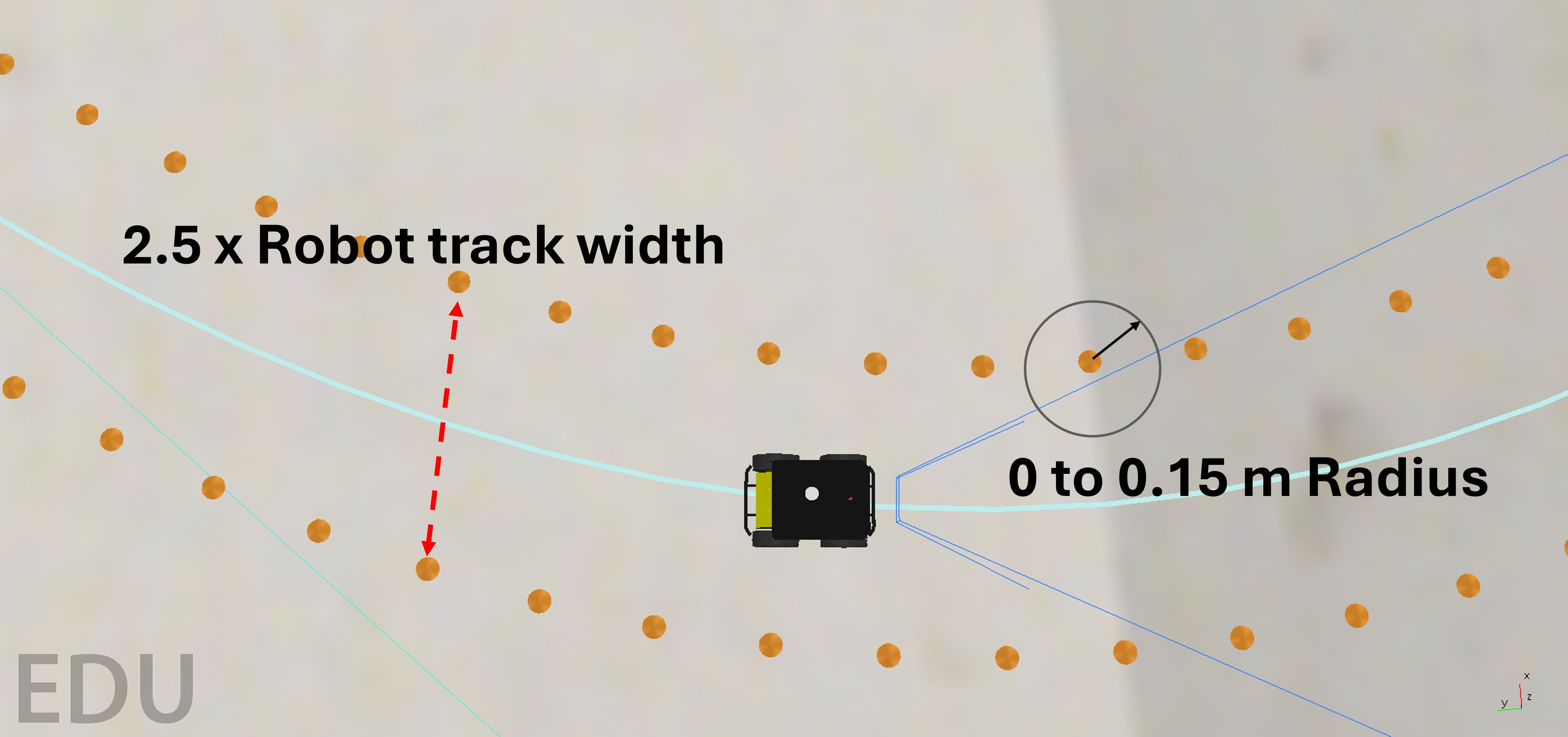}
        \vspace{0.3em}
        \small (b)
    \end{minipage}

    \caption{(a) Two arbitrary closed-loop curves. (b) The width of the lane is roughly 2.5 times the robot track width, and each cone is randomized to be located within a radius of 0.15~m from the specified center.}
    \label{fig:Learning-track}
\end{figure}

The other half of the environment apart from the robot itself is the track that is utilized for training the policy. The characterization of the similarity between the physical navigation task and its digital replica is essential not only from the point of view of creating transferable policies, but also for performance evaluation and effective recreation in the future. To achieve the same, this section tries to elucidated how paths are laid, what entails the geometric center of the path and what are the spatial limits of the path that need to be considered when transferring the learnt policy to the real robot.


\begin{enumerate}
    \item \textbf{Definition of the Closed Loop Curves:}\\
    Let two arbitrary closed loop curves (signifying left and right lanes of a track) be parameterized by \( \mathbf{C}_1(t) \) and \( \mathbf{C}_2(t) \), where \( t \in [0, 1] \) is the parameter that traces the curve:
    \[
    \mathbf{C}_1(t) = \begin{pmatrix} x_1(t) \\ y_1(t) \end{pmatrix}, \quad \mathbf{C}_2(t) = \begin{pmatrix} x_2(t) \\ y_2(t) \end{pmatrix}
    \]
    These curves are separated by a minimum distance of 1.3 meters and resemble an infinity loop as illustrated in fig.~\ref{fig:Learning-track}.

    \item \textbf{Discretization into Reference Points:}\\
    Each curve is discretized into $n$ ($n:=200$) equal parts:
    \begin{align*}    
    \mathbf{c}_{1,i} =& \mathbf{C}_1\left(\frac{i-1}{n-1}\right), \quad \mathbf{c}_{2,i} = \mathbf{C}_2\left(\frac{i-1}{n-1}\right)\\ 
    &\text{for } i = 1, 2, \dots, n
    \end{align*}
    Here, \( \mathbf{c}_{1,i} \) and \( \mathbf{c}_{2,i} \) are the discrete reference points on curves \( \mathbf{C}_1(t) \) and \( \mathbf{C}_2(t) \), respectively.

    \item \textbf{Formation of the Geometric Lane Center:}\\
    The geometric lane center \( a_i \) is defined as the midpoint of the line segment connecting \( \mathbf{c}_{1,i} \) and \( \mathbf{c}_{2,i}~\text{(as illustrated in figure~\ref{fig:Learning-track})} \):
    \[
    a_i = \frac{1}{2} \left(\mathbf{c}_{1,i} + \mathbf{c}_{2,i}\right) = \frac{1}{2} \begin{pmatrix} x_{1,i} + x_{2,i} \\ y_{1,i} + y_{2,i} \end{pmatrix}
    \]

    \item \textbf{Randomization of Cone Positions:}\\
    Cones are placed on each reference point \( \mathbf{c}_{k,i} \). Each cone is then randomly perturbed within a radius \( r = 0.15 \) meters:
    \[
    \mathbf{c}_{k,i}' = \mathbf{c}_{k,i} + r \cdot \begin{pmatrix} \cos(\beta_{k,i}) \\ \sin(\beta_{k,i}) \end{pmatrix}
    \]
    where \( \beta_{k,i} \) is a uniformly distributed random variable in \( [0, 2\pi] \), and \( k \in \{1, 2\}~\text{(as illustrated in figure~\ref{fig:Learning-track})}\).

\end{enumerate}

\subsection{Learning approach}

As emphasized in the literature, utilizing DRL for lane following is not a novel idea and has been explored within simulation settings~\cite{Chen2019,Liang2018,Toromanoff2020,Jaritz2018}. A commonality among all the works is similarity in the problem formulation method which is derived from the classical lane centering approach. The typical vision based lane centering approach involves approximating the lane center as image centroid (along the image's horizontal axis) using  image processing methods and generating control commands (steering or yaw rate) such that this image centroid is moved to the camera center. When expressed as a learning problem, DRL agent tries to learn the control actions subject to minimization of this image centering error. Unfortunately, minimizing image center error does not ensure that the robot's chassis center also aligns with the lane center. Thus, when tried to scale beyond small scaled - low speed - wide road conditions, the image centroid guided ($\pi_{ICG}$) tracking approach towards learning lane keeping falls short of optimal performance. To this end, a waypoint guided approach ($\pi_{W_{p}G}$) to learning lane keeping is proposed in this work. 
~
\subsubsection{Waypoint guided learning}~\label{subsec:guided_learning}
The definition of what is constituted as the geometric lane centre has been discussed in section~\ref{subsec:track}, and the derived waypoints points ($a_{i}$) serve as a starting point to setup the path tracking learning problem. A typical path tracking problem demands access to \textit{a reference point} at any given instance that guides the robot's motion. Identifying this reference point can be achieved by either (a) parameterizing the available waypoints in time and choosing the reference point depending on for how \textit{much time} the robot has traveled (time parameterized trajectory) , or, (b) parameterizing the available waypoints in distance and choosing the reference point depending on how \textit{much distance} the robot has traveled (arc length parameterized path). While the former is straight forward to implement, it introduces an additional velocity constraint which may require frequent re-planning for varying tracking velocities. To avoid this, the arc length path parameterization approach has been adopted for this learning formulation.

The reference points obtained from the geometric center identification process described in Section~\ref{subsec:track} are used to generate \( j \) sets of waypoints, each consisting of \( n \) points. Two operations are performed on the original waypoints to create these sets:

\begin{enumerate}
    \item \textbf{Shifting}: Half of the sets (\( \frac{j}{2} \)) are generated by circularly shifting the original waypoints. Specifically, the first waypoint in each set is shifted by \( mw \) positions, where \( m = \left\lfloor \frac{j-1}{2} \right\rfloor \) and \( w \) is a shifting constant (with \( w = 40 \)). Here, \( \lfloor \cdot \rfloor \) denotes the floor function.

    \item \textbf{Reversing}: For the remaining half of the sets, the order of each of the previously generated \( \frac{j}{2} \) sets is reversed to create new sets.
\end{enumerate}

As a result, this process produces a total of \( j \) sets of waypoints.

\noindent
For \( j = 1, 2, \dots, 10 \), we define the set of matrices \( \mathbf{A}_{j} \) as:
\begin{align*}
\mathbf{A}_{j} &= \left\{
\begin{bmatrix}
    a_{1,j} := \{x_{1 + mw} ,~ y_{1 + mw}\} \\
    a_{2,j} := \{x_{2 + mw} ,~ y_{2 + mw}\} \\
    \vdots \\
    a_{n,j} := \{x_{n + mw} ,~ y_{n + mw}\}
\end{bmatrix}_{j}
\right\}
\end{align*}
\newline
~
For any \( j \)-th set, the path between consecutive waypoints \( a_{i,j} \) and \( a_{i+1,j} \) is represented by a clothoid curve \( \mathcal{C}_{i,j}(s) \), where \( s \) is the arc length parameter. Each clothoid curve is defined as:
\[
\mathcal{C}_{i,j}(s) = \begin{pmatrix} x(s) :=\int_0^s \cos\left(\frac{\pi u^2}{2}\right) \, du  \\ y(s) :=\int_0^s \sin\left(\frac{\pi u^2}{2}\right) \, du \end{pmatrix}, \quad s \in [0, L_{i,j}]
\]
where \( L_{i,j} \) is the arc length of the curve \( \mathcal{C}_{i,j} \) from \( p_{i,j} \) to \( p_{i+1,j} \). The entire path \( \mathcal{P}_{j}(s) \) is formed by concatenating the \( n-1 \) clothoid curves:

\[
\mathcal{P}_{j}(S) = \sum_{i=1}^{n-1} \mathcal{C}_{i,j}
\]

where $S$ is the cumulative arc length (sum of all the individual $n-1$ clothoid arc lengths, $s$) along the path \( \mathcal{P}_j \), and \( L_j = \sum_{k=1}^{n-1} L_{k,j} \) represents the total arc length of the path for the \( j \)-th set. For the path \(\mathcal{P}_j\), the angle \(\theta(S)\) is defined as the angle made by the unit tangent vector, \(\hat{\mathcal{T}}_s\), at any point along the curve. A lookup table of tracking reference points \(\mathbf{X_{ref}} = [x_{r}, y_{r}, \theta_{r}]\) can then be constructed by sampling equidistant points (at an arbitary distance, $ds$, which is a design choice) along each path \(\mathcal{P}_j\) and associating them with the corresponding arc length \(S_{r}\).

For each path $P_{j}$, a table of tracking reference points $\mathbf{X_{ref}}$ is given as :

\[
\mathbf{X_{ref}}_{j} = 
\begin{bmatrix}
    S_{r,0} := {0} & x_{r,1} & y_{r,1} & \theta_{r,1} \\
    S_{r,1} := {ds} & x_{r,2} & y_{r,2} & \theta_{r,2} \\
    S_{r,2} := {2ds} & x_{r,3} & y_{r,3} & \theta_{r,3} \\
    \vdots & \vdots & \vdots  & \vdots \\
    S_{r,N} := {(N-1)ds} & x_{r,N} & y_{r,N} & \theta_{r,N} \\
\end{bmatrix}
\]

It is crucial to note that while a nominal policy can be learnt  with just one set of waypoints (instead of $j$ sets), the \textit{shifting} and \textit{reversing} steps ensure that an unbiased learning by creating variations of the learning track.

~
\subsubsection{Reward function}~\label{subsec:Reward}

The reward function within the context of a DRL problem is the mathematical realization of the per step learning objective(s), which the DRL agent tries to maximize during the learning process. The function can be composed of both the robot states and control actions and is expressed as a single scalar objective composed of different sub-objectives. Within the scope of this work, the reward function, $\mathbf{R}$ is formulated as :

\begin{align}\label{eq:RewardFunction}
\mathbf{R} = &\ (1 - e_{x})^{2} + (1 - e_{\theta})^{2} + (1 - e_{V})^{2} \notag \\
        &\ + (1 - e_{a_{1}})^{2} + (1 - e_{a_{2}})^{2}
\end{align}

Where,
\begin{itemize}
    \item \(e_{x}\): $ \| \mathbf{X}_{ref}, \mathbf{X}\| \quad \in [0,1]$ \newline
    is the euclidean distance between the reference waypoint position $\mathbf{X}_{ref,x,y}$ and the robot position, $\mathbf{X}_{x,y}$ (both in inertial frames) at any learning step.
    \item \(e_{\theta}\): $ 1- \mathbf{q}_{ref} \cdot \mathbf{q} \quad \in [0,1] $ \newline
    is the error between the reference orientation $\mathbf{X}_{ref,\theta}$ and the robot's orientation $\mathbf{X}_{\theta}$ (both in inertial frames) computed as the quaternion ($\mathbf{q}$) similarity error by expressing the reference and the robot orientations as quaternions.
    \item \(e_{V}\):  $ \| {}^{B}V_{d}, {}^{B}V\| \quad \in [0,1]$ \newline
    is the euclidean error between the desired body velocity ${}^{B}V_{d}$ (user specified) and robot's realized linear velocity, ${}^{B}V$, both in the body frame.
    \item $e_{a_{1}} : |{a_{1}}| \in [0,1]~\&~e_{a_{2}}: |{a_{2}}| \in [0,1]$\newline
    are penalties on the absolute values of the actions provide by the agent for minimizing energy consumption. It is important to note that introduction of these penalties resulted in smoother robot motions (significantly less chatter in control actions).
\end{itemize}

The choice of these parameters come from an intuitive understanding of path and trajectory tracking objectives that are captured in the literature. Formulations typically capture utilizing reference path, pose and velocity to guide the robot behavior. Penalizing actions is the popular approach in optimal control literature introducing energy optimality and often action smoothening to avoid jerky behavior.

All of the reward parameters are normalized between $0$ to $1$ as an attempt to to provide equal emphasis on each of the learning objective. Weighted combinations of reward objectives to modulate learning behaviors is an active research area but beyond the scope of this work~\cite{salvi2022stabilization}.

While the desired robot velocity, $V_{d}$ is provided at the beginning of the training process and stays constant throughout, the desired pose $\mathbf{X}_{ref}$ at every training step is provided by utilizing the robot's location and the path specific lookup tables, $\mathbf{X}_{ref,j}$, by the index identifying routine as illustrated in the algorithm~\ref{alg:wp_selection}. A path \( \mathcal{P}_r \) is randomly selected from the set of precomputed paths \( \{\mathcal{P}_1, \mathcal{P}_2, \dots, \mathcal{P}_{j}\} \), with \( r \) denoting the index of the selected path. The algorithm then computes the tracking errors \( e_x \) and \( e_\theta \), and uses them to compute the reward for reinforcement learning. The training is conducted for a total of $ t_{\text{max}}$ steps while being reset at the end of each episode. Termination of each episode is defined when: (a) robot translation error ($e_{x}$) is greater than or equal to $1.0$~m, or (b) robot orientation error ($e_{\theta}$) is greater than or equal to $0.1$~rad, or (c) the episode reaches a maximum number of transition steps per episode, ${t_{\text{max}}}^{ep} = 1000$.
    
\begin{algorithm}
\caption{Waypoint Selection Routine}
\label{alg:wp_selection}
\begin{algorithmic}[1]
\STATE \textbf{Input}: $ t_{\text{max}}~\text{and}~{t_{\text{max}}}^{ep} $
\STATE \textbf{While} \( t < t_{\text{max}} \) \textbf{do:}\\
\STATE \quad \textbf{While} \( t < {t_{\text{max}}}^{ep} \) \textbf{OR} \( e_{x} \leq 1.0 \) \textbf{OR} \( e_{\theta} \leq 0.1 \) \textbf{do:}\\
\STATE \quad \quad \textbf{if}~$t == 0$~\textbf{do:}
\STATE \quad \quad \quad \textbf{Initialize}: Select a random path\\
\quad \quad \quad \( \mathcal{P}_r = \{( S_{r,i}, x_{r,i}, y_{r,i}, \theta_{r,i})\}_{i=1}^{N+1},~\forall \mathcal{P}_r \in \mathcal{P}_{j} \)

\STATE \quad \quad \quad Set initial pose:
\[
\mathbf{X}_0 \gets \begin{bmatrix}
x_{r,1} \\
y_{r,1} \\
\theta_{r,1}
\end{bmatrix}, \quad s_0 \gets 0
\]
\STATE \quad \quad \textbf{else:} pass

    \quad \quad \textbf{State Transition}
    \STATE \quad \quad  \(\mathbf{s}_{t+1} \gets \text{Simulate}(\mathbf{s}_t, \mathbf{a}_t, \Delta t)\)\\
    
     \quad \quad \textbf{Arc Length Calculation}:
    \STATE \quad \quad \(\Delta S_t \gets \sqrt{(x_{t+1} - x_t)^2 + (y_{t+1} - y_t)^2}\)
    \STATE \quad \quad \(S_{t+1} \gets S_t + \Delta S_t\) \\
    
    \quad \quad \textbf{Index Identification}:
    \STATE \quad \quad \(i^* \gets \arg \min_{i} |S_{r,i} - S_{t+1}|\)\\
    
    \quad \quad \textbf{Position Error}:
    \STATE \quad \quad \(e_{x} \gets \sqrt{(x_{t+1} - x_{r,i^*})^2 + (y_{t+1} - y_{r,i^*})^2}\)\\
    
    \quad \quad \textbf{Orientation Error}:
    \STATE \quad \quad $\mathbf{q}_{t+1} \gets \text{Quaternion}(\theta_{t+1})$
    \STATE \quad \quad $\mathbf{q}_{r,i^*} \gets \text{Quaternion}(\theta_{r,i^*})$\\
    \STATE \quad \quad $ e_\theta \gets 1 - (\mathbf{q}_{t+1} \cdot \mathbf{q}_{r,i^*})$
    
    \STATE \quad \quad \textbf{Increment}: Increment the step counter \( t \gets t + 1 \)

\STATE \quad \textbf{End While}
\STATE \textbf{End While}
\end{algorithmic}
\end{algorithm}

\subsection{Learning hyperparameters}

All policies were trained on a server equipped with 28 CPU cores, 125 GB of RAM, and 2 NVIDIA P100 GPUs. Table~\ref{tab:hyperparameters} enlists the hyperparameters utilized for training all the policies and indicates the learning statistics for the training convergences. It is valuable to note that while visually stable convergence was witnessed at around $5e6$ timesteps, such policies did not transfer well to real world. Specifically these policies would fail on sharp curves or realized an intermittent start-stop motion. It was only after training for a minimum of $1.5e7$ steps was a stable deployment achieved. 

\begin{table}
    \centering 
        \centering
        \begin{tabular}{lc}
        \toprule
            \textbf{Hyperparameter} & \textbf{Value} \\
        \midrule
            Parallel Environments & 24 \\
            Training Steps & 2e7 \\
            Batch Size & 512 \\
            Clip Range & 0.1 \\
            Learning Rate & 1e-6 \\
            Training Epochs & 5 \\
            GAE & 0.98 \\
            $\gamma$ & 0.98 \\
        \end{tabular}
    
    \vspace{1em} 

        \centering
        \begin{tabular}{lcc}
        \toprule
            \textbf{Mean Reward} & \textbf{Mean steps} & \textbf{Clock time} \\
        \midrule
            1.80e3 & 690 & 2160 mins. \\
        \bottomrule
        \end{tabular}
    \caption{Training hyperparameters \& convergence statistics }
    \label{tab:hyperparameters}
\end{table}

In summary, section~\ref{sec:problem_formulation} captures the learning problem setup with an emphasis on the waypoints indexing approach for embedding the lane centre information in the DRL training process. It is also concluded that (a) the choice of training regime (end-to-end vs IK) and (b) sampling of reference waypoints ($ds$) are areas that need investigation. The experiements for validating the effectiveness of the learning process are thus split in two sections :
\begin{itemize}
    \item Simulation Experiments:
    These are simulation specific investigations where the choice of training regime and spacing of waypoints is investigated and its results discussed.
    \item Deployment Experiments:
    These are hardware deployments of the waypoints guided learning framework, its comparison with the image centroid based learning framework, and investigation of avenues for improved simulation to reality transfer.
       
\end{itemize}
\section{Experiments : Simulation}~\label{sec:experiments-sim}

This section captures the simulation studies that investigate the influence of introduction of model kinematics and the impact of waypoint spacing in the learning process The simulator based studies conclude by comparing the proposed policy with existing benchmark navigation frameworks such as Non-linear Model Predictive Control (N-MPC), Pure pursuit and Proportional-Derivative (PD) Control based center tracking. All simulations evaluations are conducted from the average of data of $100$ episodes for each test case. All evaluations are done on the same track with waypoint spacing $ds = 0.01$~m, to have a uniform evaluation environment.

\subsection{Waypoint spacing}~\label{subse:waypoint_spc}

\begin{figure*}[t]
    \centering

    \begin{minipage}{0.32\linewidth}
        \includegraphics[width=\linewidth]{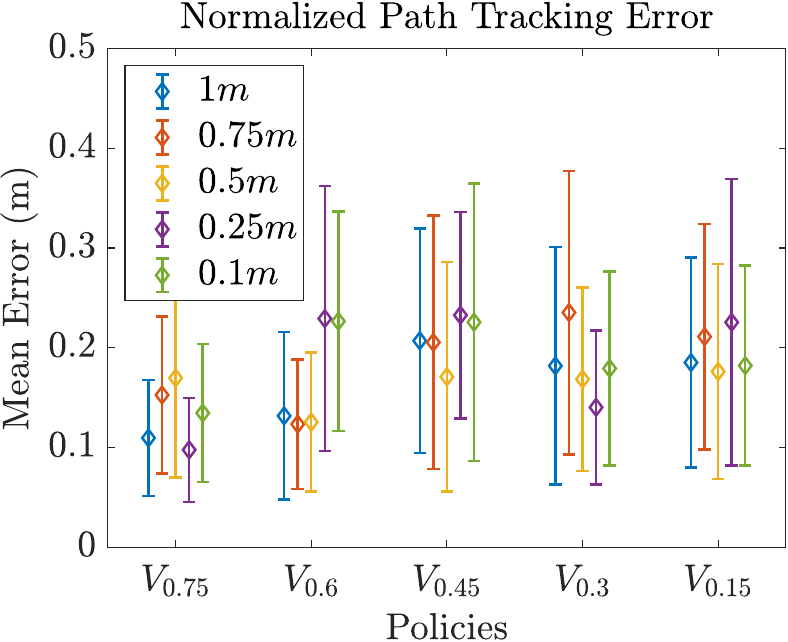}
        \vspace{0.2em}
        \centering
        \small (a)
    \end{minipage}\hfill
    \begin{minipage}{0.32\linewidth}
        \includegraphics[width=\linewidth]{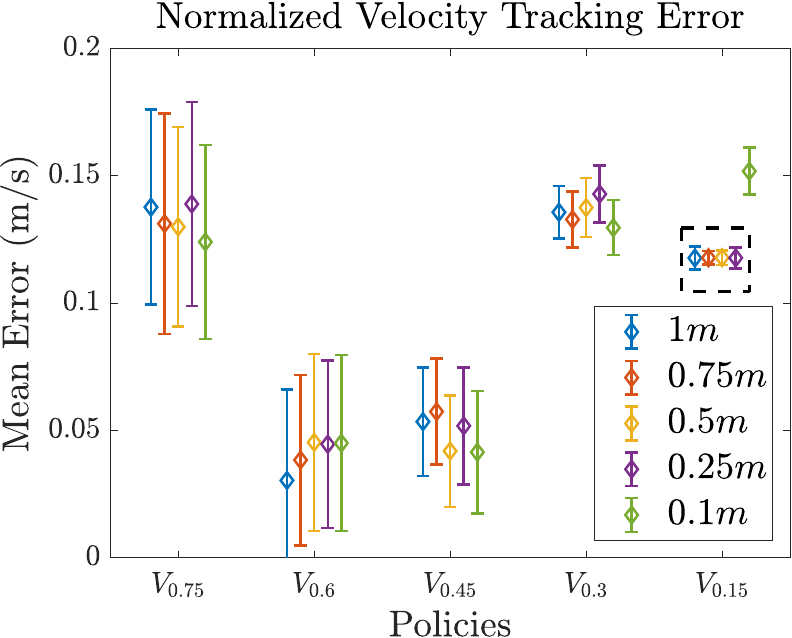}
        \vspace{0.2em}
        \centering
        \small (b)
    \end{minipage}\hfill
    \begin{minipage}{0.32\linewidth}
        \includegraphics[width=\linewidth]{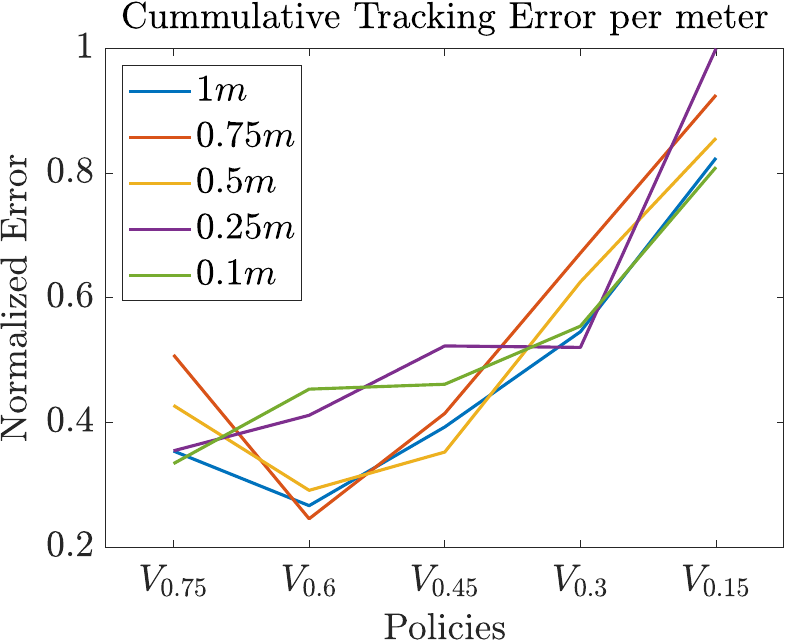}
        \vspace{0.2em}
        \centering
        \small (c)
    \end{minipage}

    \caption{Comparison of $25$ policies learnt for reference tracking velocities ${}^{B}V_{d} \in \{0.75,0.6,0.45,0.3,0.15 \}$~m/s with waypoints sampled at distances $ds \in \{1, 0.75, 0.5, 0.25, 0.1 \}$~m. (a) Path tracking error $e_{x}$. (b) Velocity tracking error $e_{V}$. (c) Distance normalized cumulative tracking error $\mathbf{N}$.}
    \label{fig:eval-waypoint-spacing}
\end{figure*}

\begin{table*}
\renewcommand{\arraystretch}{1.3} 
\centering
\begin{tabularx}{0.95\linewidth}{l*{8}{>{\centering\arraybackslash}X}}
    \toprule
    \textbf{Waypoint Spacing} & \textbf{$e_{x}$}(m) & \textbf{$e_{\theta}$}(rad) & \textbf{$e_{v}$}(m/s) & \textbf{$e_{a_{1}}$} & \textbf{$e_{a_{2}}$} & Time (ms) & $S_{term}$~(m) &\textbf{Reward} \\
    \midrule
    $ds = 1.0~\text{m}$ & 0.1630 & 0.4917e-3 & 0.0950 & 0.0036 & 0.0022 & 7.525 & 21.213 & 1.59e03  \\
    $ds = 0.25~\text{m}$ & 0.1855 & 0.6015e-3 & 0.0955 & 0.0033 & 0.0010 & 7.527 & 21.879 & 1.61e03  \\
    $ds = 0.5~\text{m}$ & 0.1621 & 0.6459e-3 & 0.0945 & 0.0064 & 0.0038 & 7.687  & 21.138 & 1.71e03  \\
    $ds = 0.75~\text{m}$ & 0.1850 & 0.6659e-3 & 0.0992 & 0.0054 & 0.0003 & 7.389 & 21.894 & 1.77e03  \\
    $ds = 0.1~\text{m}$ & 0.1895 & 0.5449e-3 & 0.0984 & 0.0006 & 0.0021 & 7.487 & 22.194 & 1.80e03  \\
    \bottomrule
\end{tabularx}
\caption{Tracking objectives and mean reward for policies with different waypoint spacing}
\label{tab:wps-stats}
\end{table*}

\begin{figure}[t]
    \centering
    \includegraphics[width=0.95\linewidth]{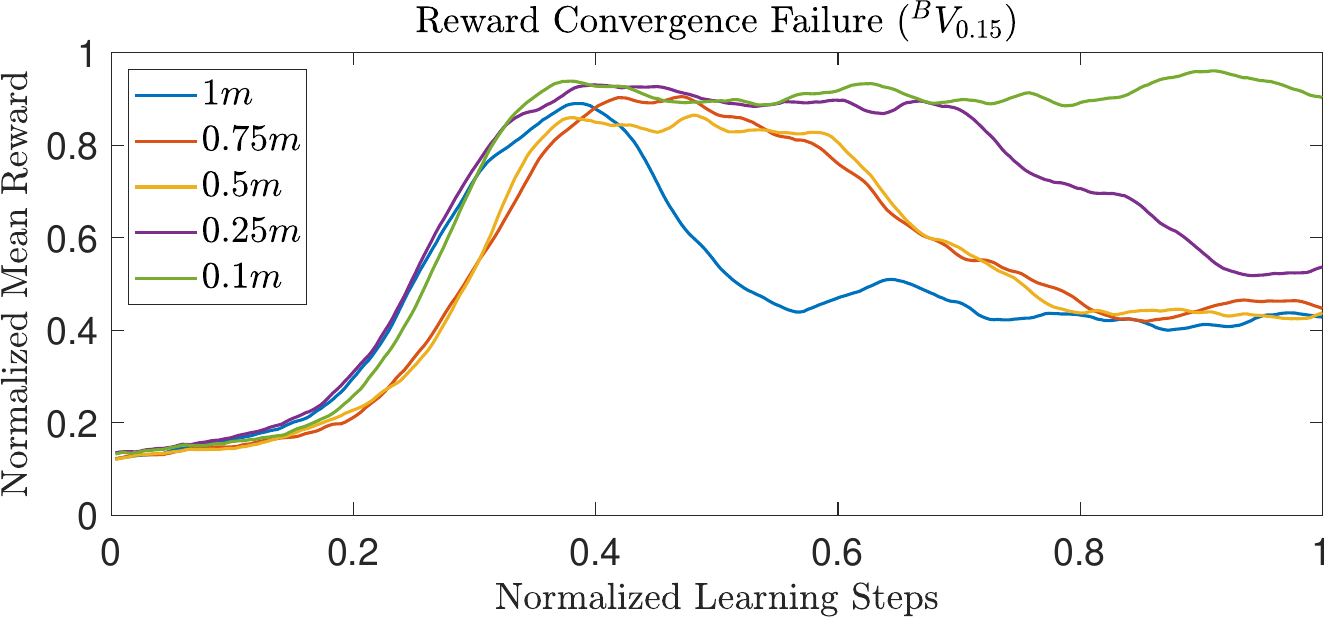}
    \caption{Reward convergence failure for reference tracking velocity of ${}^{B}V_{d} = 0.15$~m/s}
    \label{fig:RewConvFail}
\end{figure}

Section~\ref{subsec:guided_learning} introduces the idea of sampling reference points spaced at distance $ds$. Intuitively, large $ds$ leads to a coarse resolution reference path that may impact the learning performance (loss of performance on sharp corners), whereas, small $ds$ leads a fine resolution of the path which would improve performance but may increase computational load. To investigate this impact, $5$ discrete values of $ds \in \{1, 0.75, 0.5, 0.25, 0.1 \}$~m are identified and for each one of them, $5$ policies for desired body velocities ${}^{B}V_{d} \in \{0.75,0.6,0.45,0.3,0.15 \}$~m/s are trained. This generates a total of $25$ policies which can be analyzed for the learning performance.

Fig~\ref{fig:eval-waypoint-spacing} illustrates the the $25$ policies evaluated on the metrics of (a) mean of the path tracking error ($e_{x}$), (b) mean of the velocity tracking error ($e_{V}$), and, (c) distance normalized cumulative tracking error ($\mathbf{N}$):

\[ \mathbf{N} = |\frac{e_{x} + e_{V}}{S_{term}}| \]

Where, $S_{term}$~(m) is the mean distance travelled by the robot per episode before it exits the track bounds. $\mathbf{N}$ takes into account the distance traversal aspect of the learnt policy which is typically not captured in the mean of the tracking objectives. 
The results illustrate :
\begin{itemize}
    \item Contrary to intuition, finer resolution of waypoints may not always mean improved performance. The best policy performance was observed at tracking velocity of $0.6$~m/s with waypoint spacing of $1$~m.
    \item Reference tracking velocity had a significant impact (much more than waypoint spacing) on the overall policy performance. As a general trend, tracking for higher velocities yielded better results than tracking for lower velocities for all waypoint spacings.
    \item Tracking velocity of $0.15$~m/s showed the worst tracking performance (black dashed rectangle and high $\mathbf{N}$ values), which led to further analysis on that particular tracking velocity. While not evident from the plot, it is useful to know that for $0.15$~m/s, the velocity tracking error $e_v$ is due to the realized the velocity lower than the desired (${}^{B}V < {}^{B}V_{d}$ : robot moving extremely slowly) and is opposite (robot moving slightly faster than reference velocity) for all other cases.
\end{itemize}

Figure~\ref{fig:RewConvFail} illustrates the reward convergence plots for policies trained at a reference tracking velocity of \({}^{B}V_{d} = 0.15~\text{m/s}\). The results show that stable convergence to a maximum reward was achieved only at a waypoint spacing of \(ds = 0.1~\text{m}\). For all other waypoint spacings, \(ds = \{1, 0.75, 0.5, 0.25\}~\text{m}\), the policies exhibited what is commonly referred to in the literature as \textit{catastrophic forgetting}. This phenomenon occurs when the quality of the learning data degrades, leading to instability in the learning process and a failure to maintain previously achieved high rewards. 

It is important to note the trend that emerges: as the distance between waypoints increases (from \(0.25~\text{m}\) to \(1.0~\text{m}\)), the learning process begins to fail earlier in the training process. This suggests a strong correlation between decreased waypoint spacing and the likelihood of learning failure, highlighting the critical role that waypoint spacing plays in the stability and success of policy training.

Table~\ref{tab:wps-stats} provides an overview of learning process for the policies at different waypoint spacing distances, but aggregated for all the reference tracking velocities. The metric \textit{reward} is the mean episodic reward during training which can be maximum to $5000$. The readings indicate an increasing trend as the spacing distance $ds$ decreases. This should not be confused for improved performance as the nature of $(1 - e_{x})^2$ in the reward function favors finer spacing of $ds$ and thus subsequently indicate higher rewards. The `Time' column indicates the the time taken in milliseconds to execute each step and depends on the fact that the algorithm has to search through all the waypoints to find the nearest one. While in principal, decreasing $ds$ should increase the number of waypoints, it is difficult to conclude so for this particular study as the time intervals are extremely small. This effect could potentially be considerable if the training is being done in the absence of GPUs or with much finer resolution of $ds$. 

Of all, the best path tracking performance $e_x$ can be seen at $ds = 0.5$m, which interestingly shows lowest track traversal distance $S_{term}$. Conversely, the worst path tracking performance $e_x$ can be seen at $ds = 0.1$m, but shows highest distance for track traversal. This suggests that the track has sharp corners where the spacing of \( ds = 0.5 \)~m leads to failure, resulting in a lower \( S_{term} \). In contrast, at the same corners, the spacing of \( ds = 0.1 \)~m does not lead to failure, although it exhibits poor tracking performance, as indicated by the higher \( e_{x} \) but greater \( S_{term} \).

\begin{figure*}[tp]
    \centering
    \includegraphics[width=\linewidth]{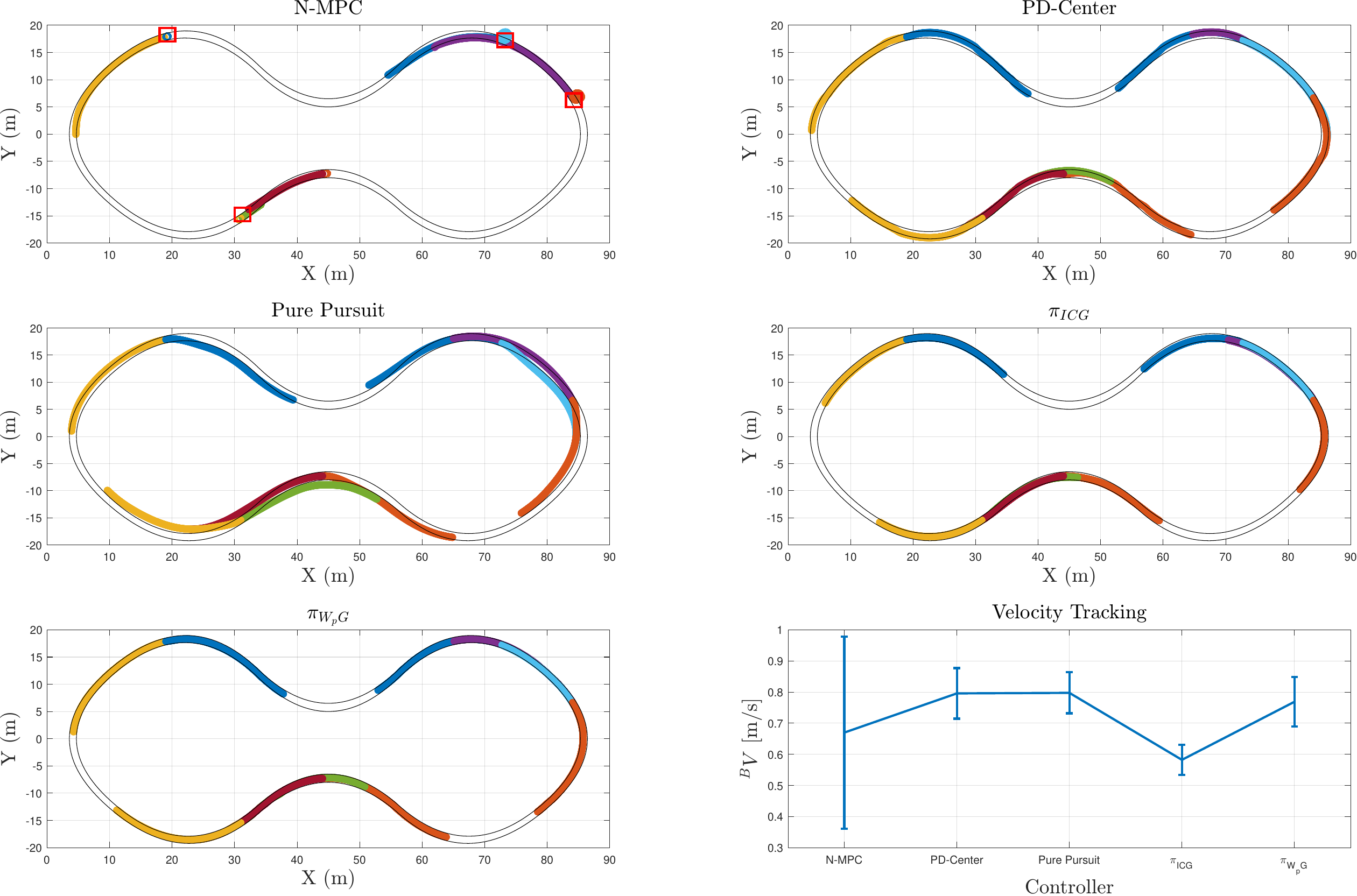}
    \caption{Comparison of paths realized by different policies on the simulation track. Figures highlight sections of the track where the controller fails (red squares for N-MPC) or sections where the robot moves beyond the track lanes. (Bottom right) Figure illustrating the body velocity (${}^{B}V$) for different control strategies (desired to be 7.5 m/s) }
    \label{fig:classical-compare-sim}
\end{figure*}

\subsection{IK vs End-to-End}

\begin{figure}
    \centering
    \includegraphics[width=0.95\linewidth]{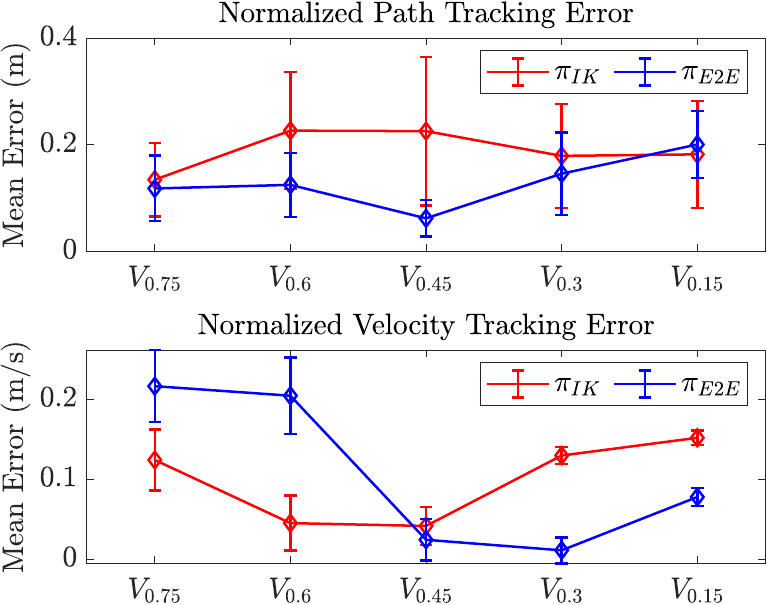}
    \caption{Comparison of reference position tracking ($e_{x}$) and reference velocity tracking ($e_{V}$) learning objectives for policies learnt with IK model ($\pi_{IK}$) and end-to-end ($\pi_{E2E}$) learning regimes for various reference velocities.}
    \label{fig:study-IKModel}
    
\end{figure}

As described in section~\ref{subsubsec:IK}, fitting an IK model is simpler than tuning the simulator to minimize the \textit{sim2real} gap. But, such a simplification can potentially reduce the policy performance, especially if the IK model is a simplified linear model as illustrated in~\ref{subsubsec:IK}. Thus, investigating the drop in learning performance by introduction of such a simplification is necessary. A total of $5$ learning policies for the two training variants ($\pi_{IK} \& \pi_{E2E}$) each were learnt for reference tracking velocities, ${}^{B}V_{d} \in \{ 0.75,0.6,0.45,0.3,0.15 \}$~m/s. The difference in the two traning regimes is the choice of the action space (body velocities vs end-to-end) as indicated in equation~\ref{eq:action-IK-E2E}. Figure~\ref{fig:study-IKModel} compares the performance of the two variants when evaluated for a total of $100$ evaluation episodes.

It is evident from fig.~\ref{fig:study-IKModel} that policy performance is much better in the case of end-to-end training regime as compared to the one one with introduction of an IK model. For both the types of learning methods, the best policy performance is obtained at ${}^{B}V_{d} = 0.45~m/s$ which lies roughly at the centre of the operating range of the robot (${}^{B}V = 0.1~\text{to}~1~m/s$). Similar to earlier discussion, for $0.15$~m/s, the velocity tracking error $e_v$ is due to the realized velocity lower than the desired velocity (${}^{B}V < {}^{B}V_{d}$) and is opposite for all other cases. It is possible that the source of this challenge could be due the fact that at much lower desired velocities, the agent is unable to resolve the conflicting objectives of tracking a low velocity ($e_{v}$) and minimizing the action effort ($e_{a1},e_{a2}$) thus resulting in a scenario where the robot barely moves. Yet, it is evident that the agent performs much better in the case of $\pi_{E2E}$ both at low speeds and overall in general. It is interesting to investigate if a similar performance could be achieved for the $\pi_{IK}$ in case a much complex IK model was used or perhaps played with weights of the reward function to emphasize more on the velocity tracking parameter, $e_v$.

\subsection{Comparison with formal methods}~\label{subsec:sim_compare_formal}

\begin{table*}
\centering
\begin{tabularx}{\textwidth}{l|*{10}{>{\centering\arraybackslash}X}}
    \hline
    & \multicolumn{2}{c}{\textbf{N-MPC}} 
    & \multicolumn{2}{c}{\textbf{PD-Center}} 
    & \multicolumn{2}{c}{\textbf{Pure Pursuit}} 
    & \multicolumn{2}{c}{$\boldsymbol{\pi_{ICG}}$} 
    & \multicolumn{2}{c}{$\boldsymbol{\pi_{W_pG}}$} \\
    \cline{2-11}
    & $e_{x}$(m) & ${}^{B}V$(m/s) 
    & $e_{x}$(m) & ${}^{B}V$(m/s) 
    & $e_{x}$(m) & ${}^{B}V$(m/s)  
    & $e_{x}$(m) & ${}^{B}V$(m/s) 
    & $e_{x}$(m) & ${}^{B}V$(m/s) \\
    \hline
    \vspace{0.5em}
    $\kappa : [0.001 \quad 0.2]$& 0.37914 & 0.84536 & 0.4056 & 0.78303 & 0.69844 & 0.78983 & 0.13372 & 0.58149 &  0.04025 & 0.75311 \\
    \vspace{0.5em}
    $\kappa : [0.2 \quad 0.4]$  & 0.18019 & 0.40175 & 0.30311  & 0.80373 & 0.86322 & 0.80113 &  0.18567 & 0.59175 & 0.036945 & 0.76651\\
    \vspace{0.5em}
    $\kappa : [0.4 \quad 0.6]$  & 0.10357 & 0.78906 & 0.18593 & 0.79615 & 0.96093 & 0.79705 & 0.15011 & 0.59128 & 0.060202 & 0.77074\\
    \vspace{0.5em}
    $\kappa : [0.6 \quad 0.8]$  & 4.1171 & 0.15054 & 0.32482 & 0.80255 & 1.2025 & 0.80299 & 0.11825 & 0.59351 & 0.10775 & 0.76791\\
    \vspace{0.5em}
    $\kappa : [0.8 \quad 0.99]$ & 1.5228 & 0.17154 & 0.33989 & 0.79001 & 1.0727 & 0.79303 & 0.083726 & 0.58688  & 0.061755 & 0.75445\\
    \hline
\end{tabularx}
\caption{Comparison of tracking error metric ($e_x$) and body velocity ${}^{B}V$ for different control policies categorized as per bins of increasing track curvature ($\kappa [1/m]$).}
\label{tab:classical_compare}
\end{table*}

Finally, the trained policy is compared with formal methods in the literature that are popular for robot navigation task. In particular, Non-linear Model Predictive control (N-MPC), Pure pursuit, Proportional Derivative control based center tracking (PD-Center), and, a DRL policy trained on centering image centroid ($\pi_{ICG}$), are chosen to compare the proposed learning framework. The first three stand as formal benchmark methods popular in the literature and $\pi_{ICG}$ is the contemporary method for learning lane keeping solutions~\cite{LK-MPC,LK-PurePursuit,Li_2023}. The detailed formulation of these frameworks has been outlined in the section~\ref{sec:Appendix}.

The performance of all the methods has been captured in table~\ref{tab:classical_compare} based on categorizing the simulation track based on path curvature ($\kappa~[1/m]$). Some critical observations as illustrated in fig.~\ref{fig:classical-compare-sim} are that there are several sections where formal controllers such as N-MPC, Pure pursuit and PD-Center tracking go out of track bounds. More critically, the N-MPC formulation is very sensitive to the input feature as it may result in a reference trajectory that may be difficult for the controller to track. Such scenarios have be high-lighted in the figure with red squares indicating the failure of the controller to come up with a feasible solution. It is critical to observe that velocity plots in figure~\ref{fig:classical-compare-sim} are show slightly higher velocities for controllers such as Pure pursuit and PD-Center tracking due to lack of feedback from the controllers. While both learning based policies perform well for staying within the track bounds, $\pi_{ICG}$ shows noticeable drop of perfromance in velocity tracking.

Table~\ref{tab:classical_compare} illustrates the path and velocity tracking performance binned according to the track curvature ($\kappa$). The table illustrates that most of the formal methods start to perform noticeably poorly as the track curvature keeps on increasing. While controllers such as Pure pursuit and PD-Center tracking come up only with the angular velocity values, the body velocity tracking performance is fairly well for these controllers. Unfortunately, N-MPC which calculates both linear velocity and angular velocity shows poor performance, which may potentially be improved by weight tuning for MPC cost function.

\section{Experiments : Deployment}~\label{sec:experiments-deploy}

\begin{figure*}
    \centering
    \includegraphics[width=0.95\linewidth]{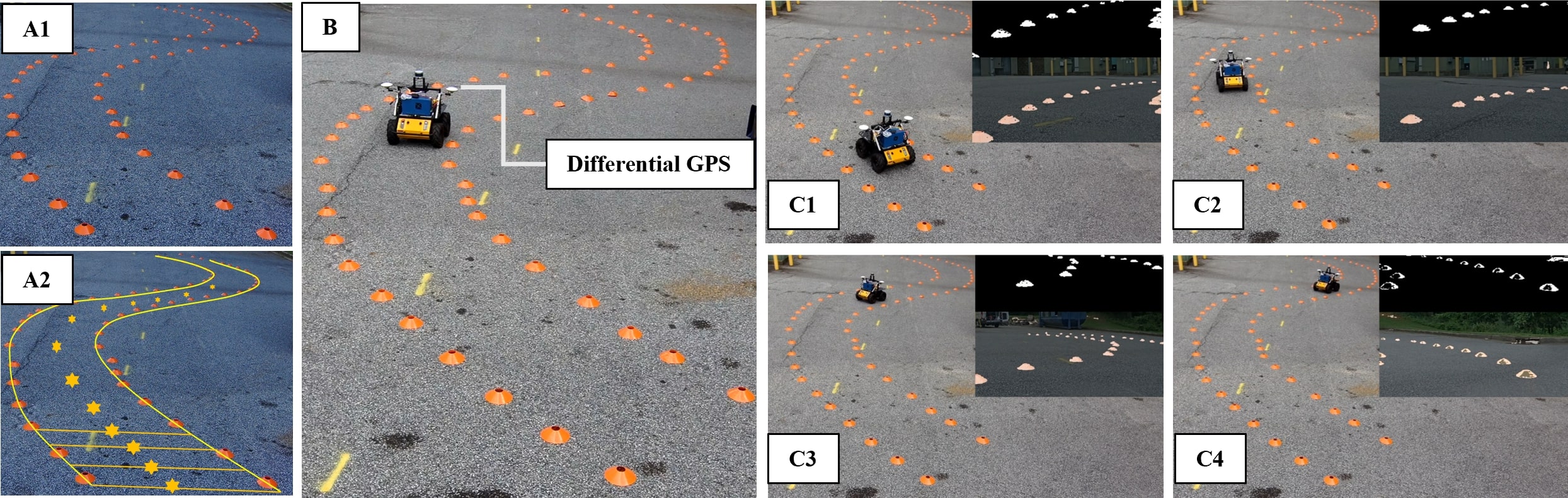}
    \caption{Overview of hardware deployments: [B] Policies evaluated for visual navigation with soccer cones as visual markers and utilizing onboard differential global positioning system (GPS) for post-processing the trajectories. [A1-A2] Identifying geometric lane center by first remote teleoperating the robot over the right and left lanes and utilizing the geometric center identification method identified in~\ref{subsec:track}. [C1-C4] Time sequence images of the robot traversing with the baseline policy with illustration of the feature inputs.}
    \label{fig:PhysicalExperiments}
\end{figure*}

\begin{figure*}
    \centering

    \begin{minipage}{0.95\linewidth}
        \centering
        \includegraphics[width=\linewidth]{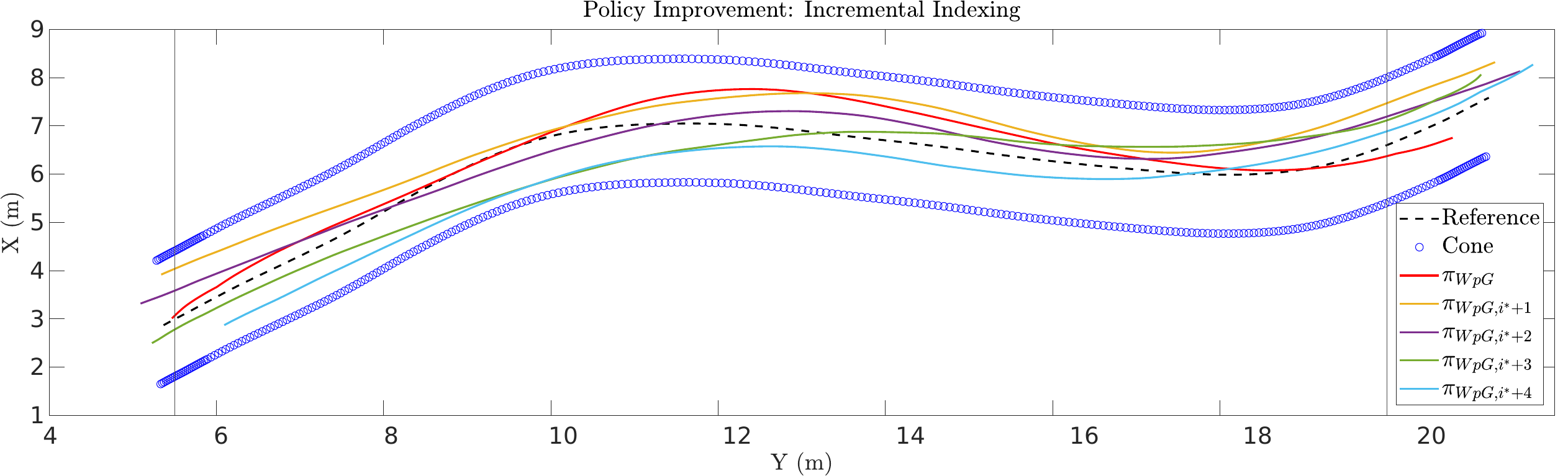}
        \vspace{0.5em}
        \small (a)
    \end{minipage}

    \vspace{1em} 

    \begin{minipage}{0.475\linewidth}
        \centering
        \includegraphics[width=\linewidth]{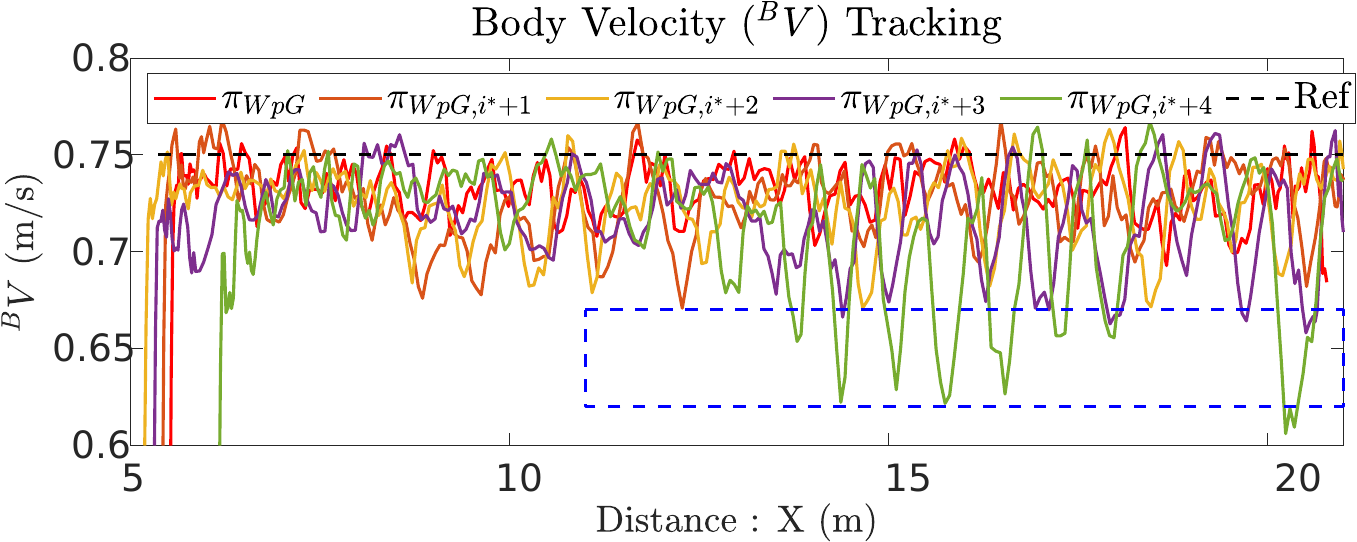}
        \vspace{0.5em}
        \small (b)
    \end{minipage}\hfill
    \begin{minipage}{0.475\linewidth}
        \centering
        \includegraphics[width=\linewidth]{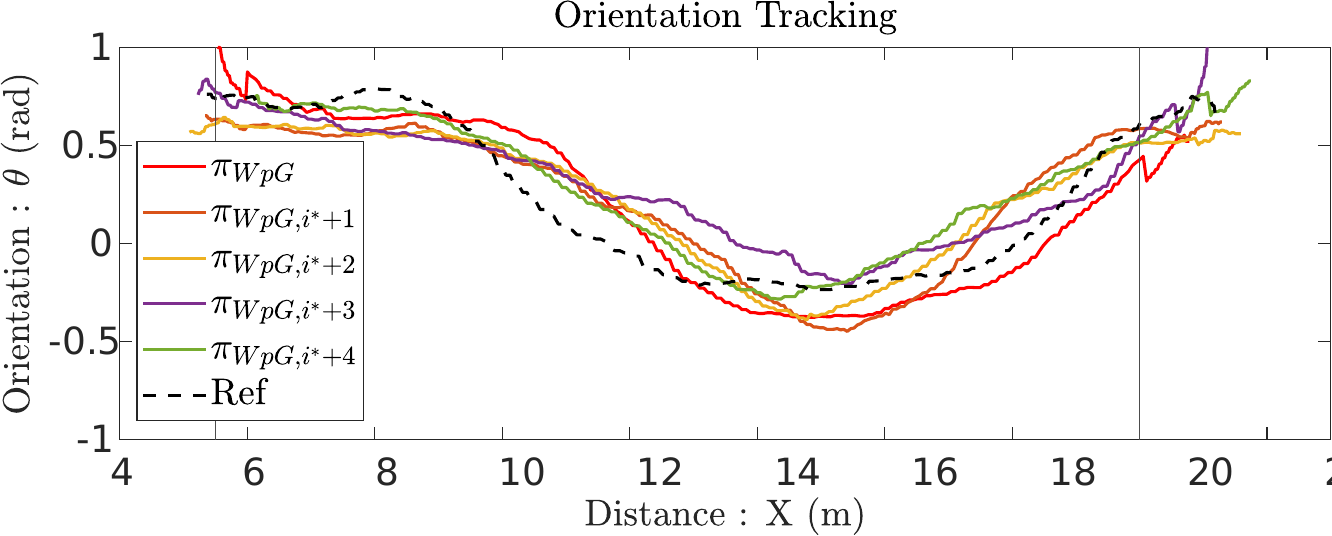}
        \vspace{0.5em}
        \small (c)
    \end{minipage}

    \caption{(a) GPS fused odometry paths realized for four different policies trained with $\alpha = \{1,2,3,4\}$. Compared to the baseline policy $\pi_{W_pG}$, all new policies except the one with $\alpha = 1$ significantly undershoot on corners. (b) Realized body velocity (${}^{B}V$) for all policies. The policy with $\alpha = 4$ exhibits erratic velocity tracking behavior, with much lower realized velocities (highlighted in the blue box). (c) Realized orientation trajectories for all policies show minimal deviation.}
    \label{fig:IdxPlots}
\end{figure*}


For all physical deployments, the policies trained with the inverse kinematics model and waypoint spacing $ds = 0.1$~m were chosen. Figure~\ref{fig:PhysicalExperiments} illustrates the process of physical validation of the control policies. The tracks are laid using orange soccer cones similar to the ones in training over a track length of roughly $140$ m and lane width roughly $1.5$ m. The robot is equipped with an onboard differential GPS sensor suit, which coupled with inertial odometery is utilized to validated the tracking performance. The robot is first teleoperated with precision over the right and left lanes to capture the location of the cones. Based on captured paths, the geometric lane center is identified by the process outlined in~\ref{subsec:track}. These steps are visually illustrated in subfigures~\ref{fig:PhysicalExperiments}-A1 and~\ref{fig:PhysicalExperiments}-A2, where it can be observed that geometric lane center is identified not necessarily by connecting the cones, but the continuous path along which the cones are placed. Similar to training, $X_{ref}$, a lookup table of distance stamped reference waypoints are identified for the policy performance evaluation.  

Several experiments where identified to evaluate the policy performance of the waypoints guided policy ($\pi_{W_pG}$) proposed in this work.
\begin{itemize}
    \item \textbf{Sim2Real improvement} : The proposed approach ($\pi_{W_pG}$) is evaluated for sim2real improvements, if any, by introducing a look-ahead constant during reference waypoint selection. 
    \item \textbf{Robustness to sensor dropout} : The proposed approach ($\pi_{W_pG}$) is evaluated for robustness when input images are received at a frequency much lower than that during the training.
    \item \textbf{Image centroid guided ($\pi_{ICG}$) vs Waypoint guided ($\pi_{W_pG}$)} : The standard approached to learning lane keeping is compared to the proposed waypoints guided policy learning.
    \item \textbf{Comparison with formal methods} : Comparing $\pi_{W_pG}$ with formal methods such as Pure pursuit and PD Error based tracking for sensitivity to input feature variability.
\end{itemize}

\subsection{Waypoint selection}~\label{subsec:waypoint_select}
\begin{table*}
\centering
\begin{tabularx}{0.95\linewidth}{l*{4}{>{\centering\arraybackslash}X}|*{3}{>{\centering\arraybackslash}X}}
    \hline
    & \multicolumn{4}{c|}{\textbf{Simulation}} & \multicolumn{3}{c}{\textbf{Reality}} \\
    & $e_{x}$(m) & $e_{\theta}$(rad) & $e_{v}$(m/s) & $S_{term}$~(m) & ${e^*}_{x}$~(m) & ${e^*}_{\theta}$(rad) & ${e^*}_{V}$~(m/s) \\
    \hline
    Baseline ($\pi_{W_{p}G}$) & 0.0856 & 3.59e-4 & $0.1230^{+}$ & 34.50 & 0.7318 & 0.0205 & $0.0235^{-}$  \\
    $\alpha=1$ & 0.1119 & 10e-4 & $0.0544^{+}$ & 27.75 & \textbf{0.4810} & \textbf{0.0120} & $0.0272^{-}$  \\
    $\alpha=2$ & 0.0911 & 5.14e-4 & $0.0311^{+}$ & 30.30 & 0.7604 & 0.0338 & $0.0288^{-}$  \\
    $\alpha=3$ & 0.1507 & 3.70e-4 & $0.0413^{+}$ & \textbf{36.00} & 1.0656 & 0.02019 & $0.0371^{-}$  \\
    $\alpha=4$ & 0.1239 & 11e-4 & $0.0903^{+}$ & 29.95 & 1.4698 & 0.2855 & $0.0470^{-}$  \\
    \hline
\end{tabularx}
\caption{Comparison of Errors in Simulation and Reality when learning introduces with the factor $\alpha:=\{1,2,3,4\}$}
\label{tab:errors_idx}
\end{table*}

\begin{figure}[t]
    \centering
    \includegraphics[width=0.95\linewidth]{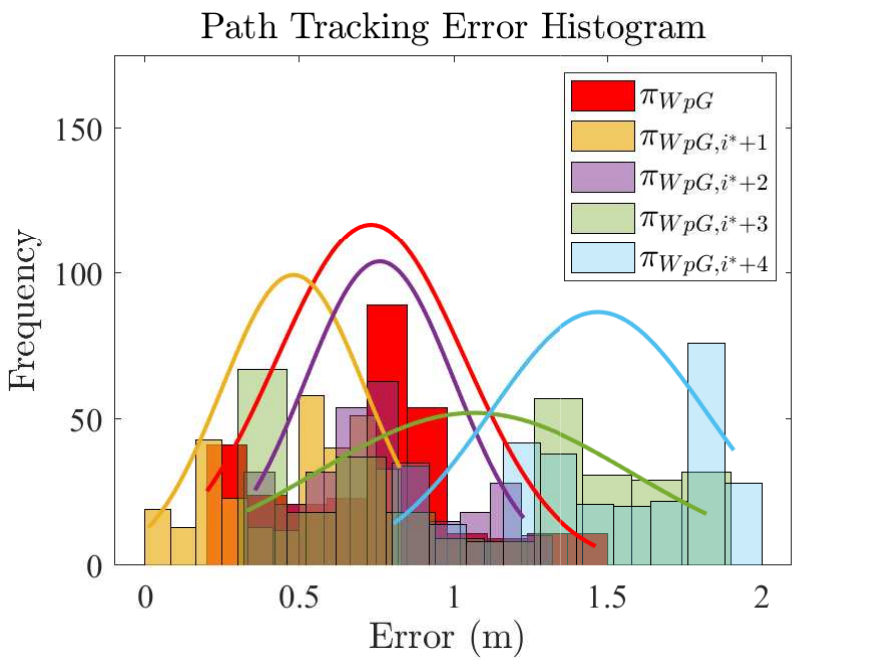}
    \caption{Path tracking error histogram for the policies with the modified waypoint selection with the bell curves representative of the individual histograms. Figure illustrates introduction of an increment in the waypoint selection with $\alpha =1$ marginally improves the sim2real transfer ( indicated by yellow bell curve moving close to zero) and additional lookahead ($\alpha>1$) again deteriorating the performance.}
    \label{fig:IdxErrHist}
\end{figure}

As discussed earlier, the inaccuracies in the simulator often lead to suboptimal performance of the polices when deployed. The most common and intuitive discrepancy is the unmodeled friction due to which the robot in reality traverses less distance as compared to the robot in the simulator. In context of this work, the arc length $S$ travelled by the robot in reality will always be slightly less than that travelled by the one in simulator for the same control inputs. To alleviate this challenge, the proposed learning approach is modified by introducing a look-ahead constant, $\alpha$, which tries to capture the sim2real discrepancy by changing the reference waypoint index $i^{*}~\text{to}~i^{*}+\alpha$. But introduction of a look-ahead distance can have a negative effect on overall policy performance as well. As researched in topics such as regulated or adaptive waypoint following methods~\cite{macenski2023regulatedpurepursuitrobot}, introduction of a look-ahead can introduce a preview of the horizon, thus improving the performance in the long run, but may lead to missing waypoints on sharp corners. Thus, finding the correct look-ahead distance that keeps the performance optimal, both in simulation and reality can be a matter of systematic investigation.

The proposed change modifies the waypoint selection routine in algorithm~\ref{alg:wp_selection} only in step 11 as follows:
\begin{align*}
i^* \gets &\arg \min_{i} |s_{r,i} - (s_{t+1})| + \alpha \\
&\forall~\alpha \in \{1,2,3,4\}
\end{align*}

The three plots is Figure~\ref{fig:IdxPlots} illustrates the GPS trajectories realized by the baseline and the modified polices after introduction of the look-ahead constant. The three figures capture the position $X$ (separated as $[x,y]~\text{and}~\theta$ ), and, body longitudinal velocity ${}^{B}V$. With visual inspection, when compared with the baseline policy ($\pi_{W_pG}$), it can be observed that he policy with $\alpha = 1$ performs similar to the baseline policy. The remaining of the policies, $\alpha \in \{2,3,4\}$ seem to undershoot on the curves. It can be also observed that the policy with $\alpha = 4$ significantly under-performs for the velocity tracking objectives which is highlighted in the blue dashed box. There isn't any significant difference in the orientation tracking objectives between all the policies. The overall behavior is expected as introduction of a look-ahead encourages the agent to realize smoother transitions, which are beneficial for straights but can be a challenge on sharp curves

Table~\ref{tab:errors_idx} captures the values between the policy performance between simulation and reality for varying $\alpha$. The path tracking error, $e_{x}$, the orinetation tracking error, $e_{\theta}$, the velocity tracking error, $~\text{and}~e_{V}$ and the traversal distance $S_{term}$ are the comparison metrics in simulation for the two policies. For the hardware evaluations, the table contains values for ${e^*}_{x},{e^*}_{\theta}~\text{and}~{e^*}_{V}$, which are similar metrics but with an $*$ indicating hardware deployment. It is valuable to note the velocity tracking performance $e_{v}$ for both the policies in simulation and reality indicating a subscript $+$ and $-$. Here $+$ indicates that the tracking error is due to realized velocity greater than the reference value and vice versa. This is an indicator that there are always losses in reality which limit the policy performance as compared to simulation

Trends indicate that addition of the look-ahead constant does tend to deteriorate the path tracking performance in simulation. But, it is valuable to note that introduction of a look-ahead distance improves the simulation traversal distance $S_{term}$ steadily, peaks for $\alpha = 3$, and then drops again. For hardawre deployment, it is observed that introduction of a small look-ahead $\alpha = 1$ improves the path and pose tracking performance. Beyond that, the performance starts to deteriorate to end up being worse for $\alpha = 4$. Figure~\ref{fig:IdxErrHist} illustrated the path tracking error histogram for the policies with modified waypoint selection. The histogram indicates the confidence on the error values indicated by recurrence of the error values which is comparable for all the policies except the one with $\alpha = 3$. This information can be of value in understanding the overall policy performance inclusive of the recurrence of the error values through the entire path. The curves on the top of the histograms indicate the approximation for the histogram bars as a Gaussian distribution. This approximations allows to compare the spread of the path tracking error values amongst all the policies and compare them with the baseline as well.

\subsection{Aynchronous input image}
Another robustness experiment in policy evaluation entails testing the policy robustness in the presence of vision sensor malfunction. In this work, sensor malfunction is characterized by asynchronous input image frequency where the fresh input image is available at a frequency much lower than for which the policy has been trained for. This is a common situation for controller deployment on edge devices which may present the control policy with processed images at a variable frequency subject to distribution of compute resources for other tasks. 

\begin{figure}
    \centering
    \includegraphics[width=0.9\linewidth]{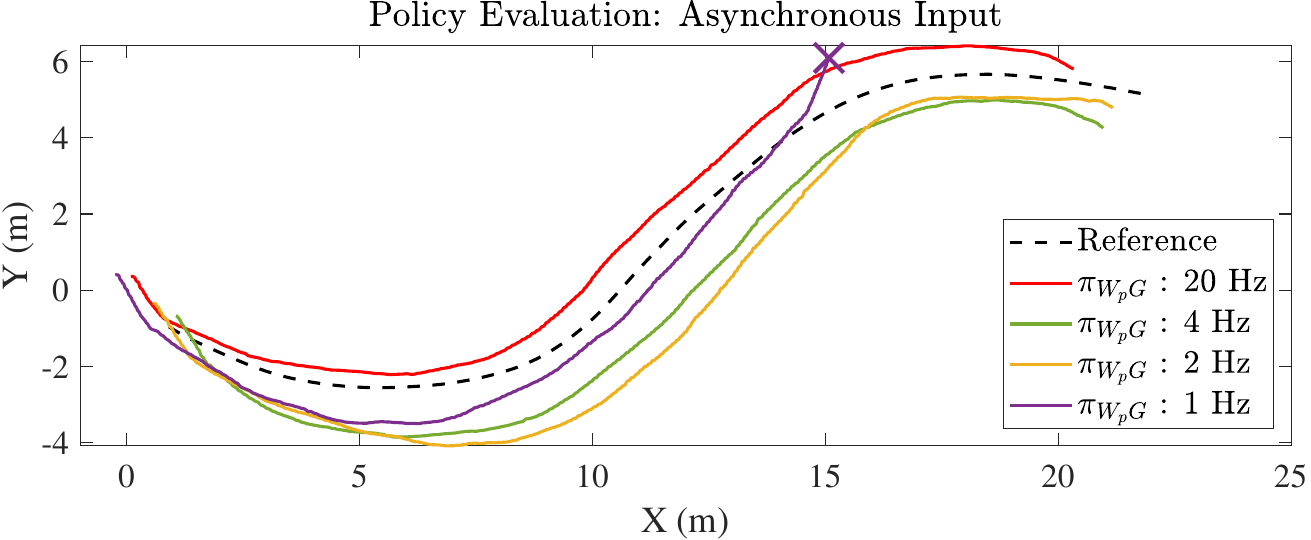}
    \caption{Motion trajectories for policy evaluation at input frequencies of 20 (Baseline), 4 ,2 and 1 Hz indicating unidirectional drift in realized motion with a track traversal failure at input frequency of 1 Hz.}
    \label{fig:deploy_async}
\end{figure}

\begin{figure}[t]
    \centering
    \includegraphics[width=0.9\linewidth]{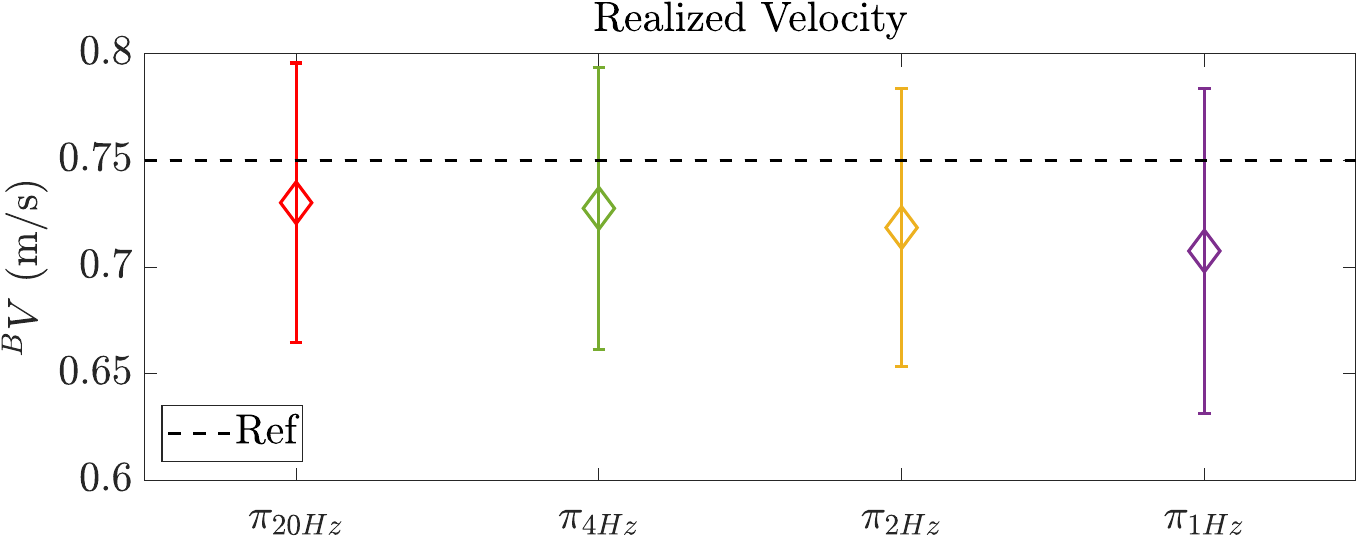}
    \caption{Trend in decreasing realized body velocity (${}^{B}V$) for policy evaluations at input frequencies of 20 (Baseline), 4 ,2 and 1 Hz}
    \label{fig:vel-async}
\end{figure}

In this work, the original policy is trained for receiving input images at $20$ Hz and has been subject to robustness testing for input frequencies of $4$ Hz, $2$ Hz and $1$ Hz. While in reality, the policy did was tested for frequencies between $20$ Hz to $5$ Hz, there wasn't any reportable discrepancy as compared to the baseline case. Noticeable changes were observed only for frequencies lower than $5$ Hz, which are included in this report. For the experiment, once a new image is acquired, the control policy keeps on receiving the same image till a fresh new image is available (which could be after $20$, $4$, $2$ or $1$ Hz depending on the experiment.)

\begin{table}
    \centering
    \begin{tabular}{cccc}
        Input Frequency & ${e^*}_{x}$~(m) & ${e^*}_{\theta}$(rad) & ${e^*}_{V}$~(m/s) \\
        \hline
        20 Hz & 0.9889 & 0.7767 & $0.0198^{-}$ \\
        4 Hz & 1.2984 & 0.2251 & $0.0225^{-}$ \\
        2 Hz & 1.2218 & 0.3677 & $0.0315^{-}$ \\
        1 Hz & -- & -- & $0.0430^{-}$ \\
    \end{tabular}
    \caption{Performance drop in path, pose and velocity tracking performance when input frequency constricted to 4, 2 and 1 Hz.}
    \label{tab:error_async}
\end{table}

Figure~\ref{fig:deploy_async} captures the motion trajectories for polices evaluated for asynchronous input images. The primary observation is that while policies constricted with input frequencies at 4 Hz and 2 Hz work, the one at 1 Hz fails around the second corner of the track. It is also notable to observe that while the baseline case (20 Hz) shows a bias on the left side of the reference path (when visualized in robot reference frame), all the policies with constricted input frequencies show a bias on the right side of the reference track. A potential explanation for this could be the  fact that the policy does not provide adequate turning motion to the robot on the first turn, which accumulates and stays unchanged for the rest of the track. On the next turn, the policy behaves similarly, but now in the opposite direction, which makes the trajectories appear converging to the reference. Figure ~\ref{fig:vel-async} indicates the policy performance for the velocity tracking objective by illustrating the realized body velocities ${}^{B}V$ for all the policies, for all of which, the  body velocity (${}^{B}V_{d}$) is $0.75$~m/s. Almost a non-linear drop in performance can be observed with lowering the input image frequency which aligns with the expected behavior. 

Table~\ref{tab:error_async} captures the overall performance of all the tracking metrics. While there is a general lowering in tracking performance, deducing the exact numerical trend may require significantly more data and potentially analyzing with a much finer granularity of input image frequencies (spacing at 0.5 to 0.1 Hz).

\begin{figure*}[tp]
    \centering

    \begin{minipage}{0.95\linewidth}
        \centering
        \includegraphics[width=\linewidth]{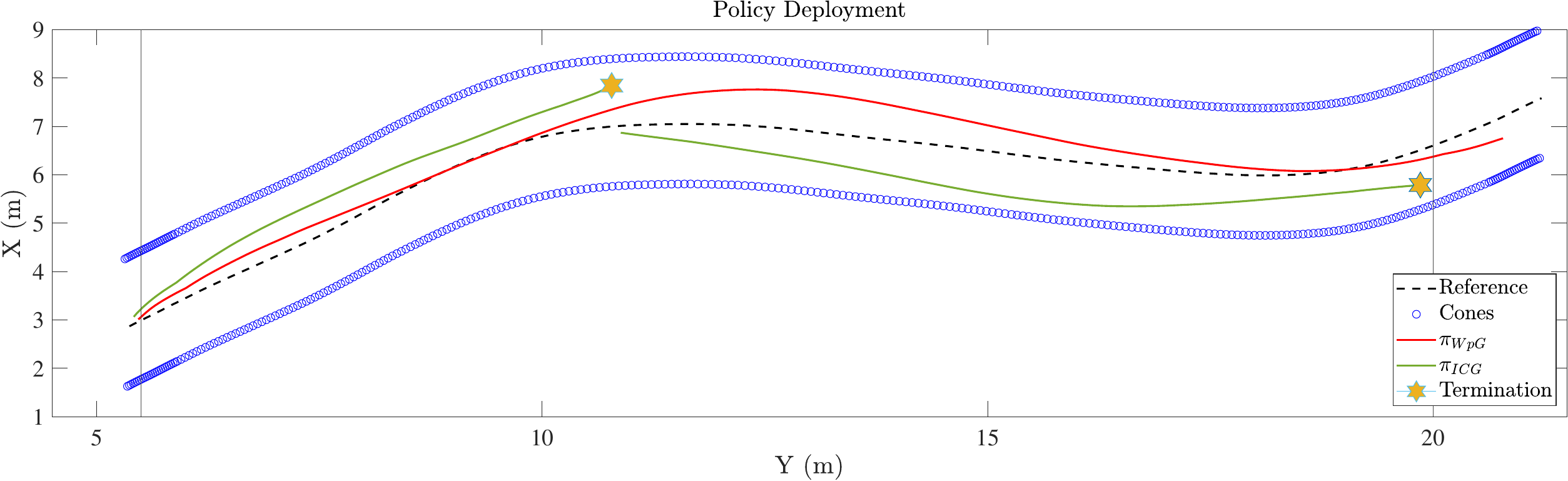}
        \vspace{0.5em}
        \small (a)
    \end{minipage}

    \vspace{1em}

    \begin{minipage}{0.475\linewidth}
        \centering
        \includegraphics[width=\linewidth]{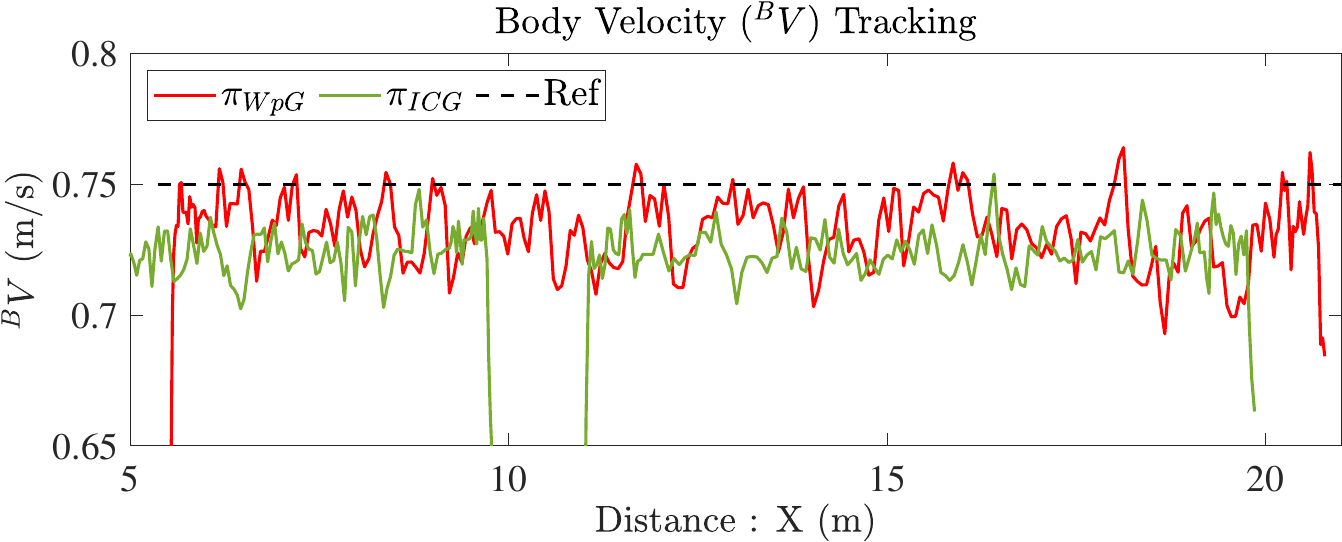}
        \vspace{0.5em}
        \small (b)
    \end{minipage}\hfill
    \begin{minipage}{0.475\linewidth}
        \centering
        \includegraphics[width=\linewidth]{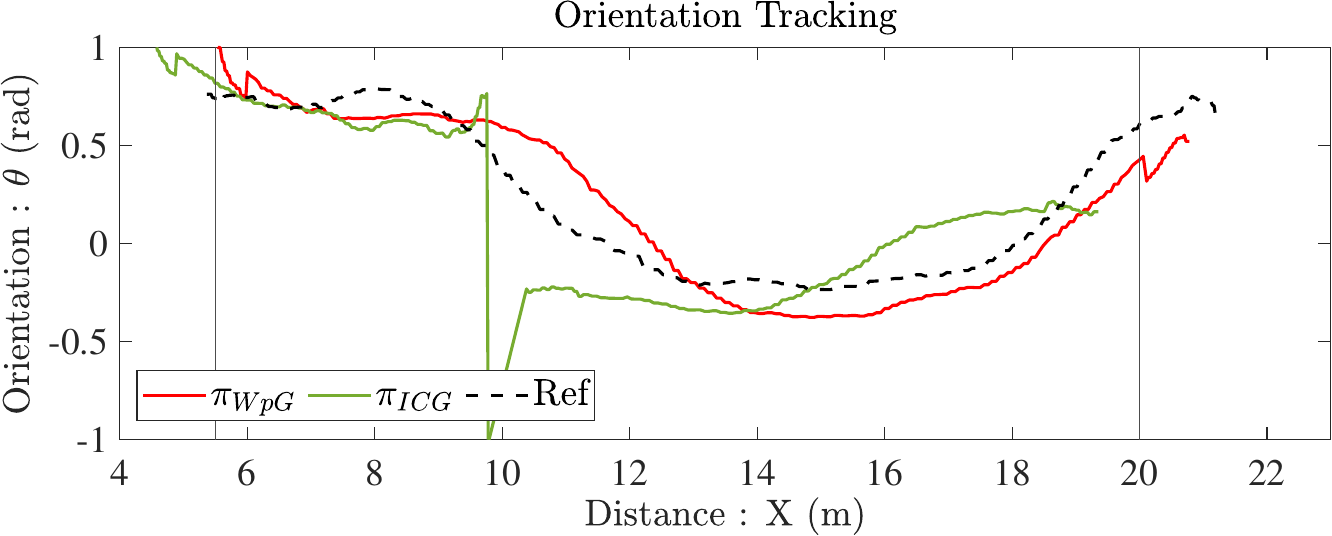}
        \vspace{0.5em}
        \small (c)
    \end{minipage}

    \caption{(a) GPS fused odometry paths realized for the waypoint-guided ($\pi_{W_pG}$) and image-centroid-guided ($\pi_{ICG}$) policies. The black vertical lines indicate the evaluation window. The $\pi_{ICG}$ policy fails around sharp corners. (b) Realized body velocity (${}^{B}V$) for both policies. (c) Realized orientation trajectories for both policies.}
    \label{fig:BslnPlots}
\end{figure*}


\subsection{Image centroid guided vs Waypoint guided}
\begin{figure}
    \centering
    \includegraphics[width=0.95\linewidth]{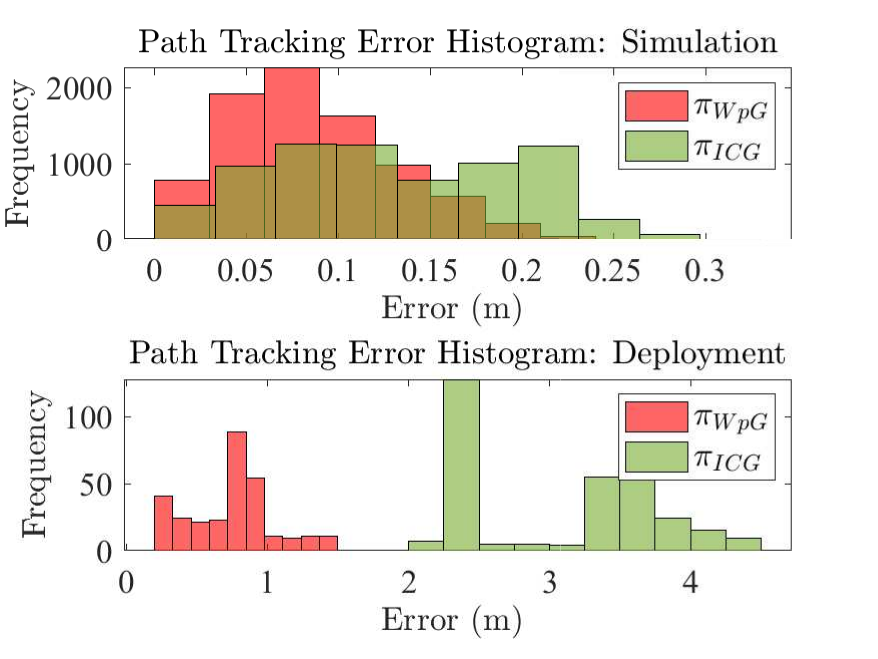}
    \caption{Path tracking error histogram for the policies $\pi_{ICG}~\text{and}~\pi_{WPG}$ in simulation and deployment.}
    \label{fig:BslnErrHist}
\end{figure}

\renewcommand{\arraystretch}{1.5} 
\begin{table*}
\centering
\begin{tabularx}{0.95\linewidth}{l*{4}{>{\centering\arraybackslash}X}|*{3}{>{\centering\arraybackslash}X}}
    \hline
    & \multicolumn{4}{c|}{\textbf{Simulation}} & \multicolumn{3}{c}{\textbf{Reality}} \\
    & $e_{x}$~(m) & $e_{\theta}$~(rad) & $e_{V}$~(m/s) & $S_{term}$~(m) & ${e^*}_{x}$~(m) & ${e^*}_{\theta}$~(rad) & ${e^*}_{V}$~(m/s) \\
    \hline
    $\pi_{W_{p}G}$ & 0.076 & 3.59e-4 &$0.0860^{+}$  & 34.5  & 0.7318 & 0.0205 & $0.0235^{-}$ \\
    $\pi_{ICG}$ & 0.131 & 7.36e-4 & $0.15^{-}$ &  29.95 & -- & -- & $0.0356^{-}$  \\
    \hline
\end{tabularx}
\caption{Comparison of Errors in Simulation and Reality for policies $\pi_{ICG}~\text{and}~\pi_{W_pG}$}
\label{tab:Bsln-errors}
\end{table*}


\begin{figure*}[t]
    \centering
    \includegraphics[width=0.95\linewidth]{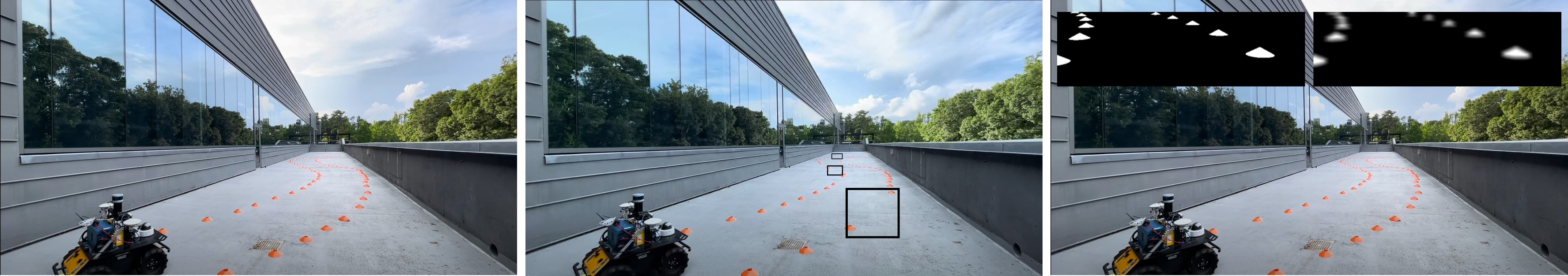}
    \par\vspace{0.8cm}  
    \includegraphics[width=0.95\linewidth]{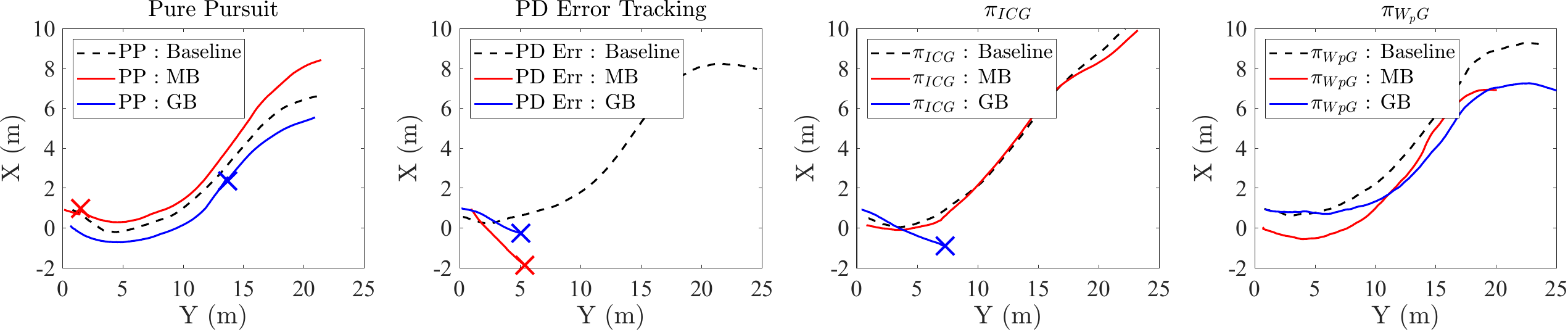}
    \caption{(Top Row) Evaluation scenarios to compare the drop in performance for lane keeping controllers. Form left to right : (1) Baseline,(2) missing sections of lane boundary, (3) Blurry lens or out of focus vision data. (Bottom row) Visualization of trajectories for the four controllers (Pure pursuit, PD-Center tracking, $\pi_{ICG}$ and $\pi_{W_pG}$) with respect to their own trajectories as baselines under ideal scenarios. Cross indicates failure of the controller or trajectory termination.}
    \label{fig:eval_scenarios}
\end{figure*}


\begin{figure*}
    \centering
    \includegraphics[width=1\linewidth]{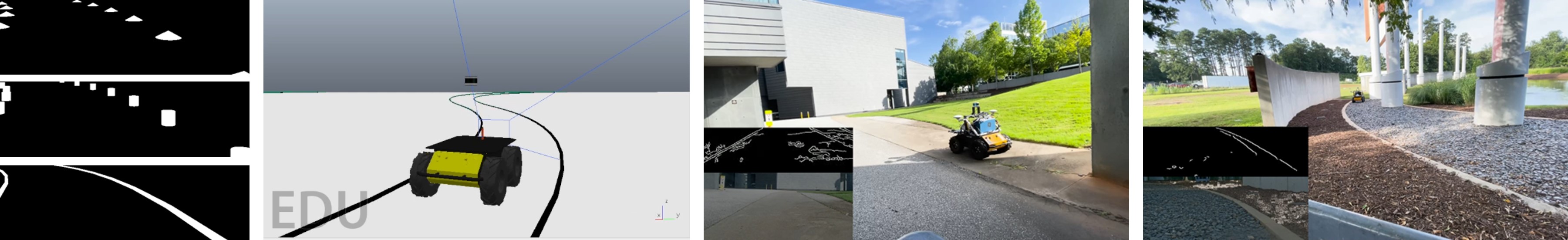}
    \caption{Zero-shot policy evaluation on out of training distribution visual features such as cylinders and solid lanes both in simulation and reality. Visuals and videos indicate policy is able to perform without any post training on visual markers available by edge detection frameworks such as Canny edge detection.}
    \label{fig:out-of-training}
\end{figure*}

As briefly discussed in Section~\ref{subsec:guided_learning}, the current state-of-the-art in learning lane centering involves training policies to manipulate the image-derived lane center. To establish a comparison, we train an image centroid-guided policy (\(\pi_{ICG}\)) on the same training track and evaluate it against the proposed waypoints-guided learning approach (\(\pi_{W_{p}G}\)) for a one to one evaluation.

Figure~\ref{fig:BslnPlots} illustrates the GPS trajectories realized by the two policies $\pi_{ICG}~\text{and}~\pi_{W_pG}$. It can be observed that $\pi_{ICG}$ terminates intermittently on sharper corners of the track. The failed policy is re-oriented to the track center and run again, only to terminate at the next sharp corner. The orientation tracking performance also indicates that it is poor when compared to the $\pi_{W_pG}$ policy. While the velocity tracking performance is more or less comparable, it is slightly better in the case of $\pi_{W_pG}$. Figure~\ref{fig:BslnErrHist} compares the two policies for the path tracking error metric in simulation and deployment. The histograms indicate that for both policies the \textit{deployment} performance deteriorates as compared to the \textit{simulation} performance. Among the two policies, $\pi_{ICG}$ deteriorates more in deployment as compared to $\pi_{W_pG}$ indicating that $\pi_{W_pG}$ is better for sim2real transfer. 


Table~\ref{tab:Bsln-errors} compares the three tracking objectives in simulation and deployment.The metrics indicate that the overall performance for the $\pi_{W_pG}$ policy is better both in simulation and reality. The proposed policy $\pi_{W_pG}$ is much better on all counts in simulation, which is an indicator that regardless of sim2real transfer, the learning performance of $\pi_{W_pG}$ is much better for the lane following task.

Because calculating the mean error is difficult due to policy failure, figure~\ref{fig:BslnErrHist} computes the histogram of the path tracking error for a clearer comparison. The histogram illustrates that overall the proposed policy ($\pi_{W_pG}$) works much better when deployed on actual hardware as compared to the image centroid guided policy ($\pi_{ICG}$). Comparing the ratios for the mean of the histogram (which is roughly at 0.73 m for $\pi_{W_pG}$ and roughly at 2.5 m for $\pi_{ICG}$) to the simulation path tracking performance ($e_x$), it can be observed that the policy deteriorates much more in the case of $\pi_{ICG}$ when deployed on hardware. One reason for this can potentially be that the error dynamics for $e_c$ as compared to $[e_x e_{\theta}]$ are much more sensitive to robot control inputs, making the policy learnt with the former less robust in sim2real transfer. The performance can further deteriorate in the presence of poorly calibrated vision sensor in the simulator, especially in the presence of weak image thresholding frameworks for marker detection.

While the policy $\pi_{ICG}$ failed to complete this track, it is essential to note that the policy could track less sharper curves. Introducing weight on the penalty for centroid tracking ($(1-e_c)^2$) has also shown to modulate the policy performance, a thorough investigation of which could be useful, but is beyond the scope of this research.

\subsection{Comparison with formal methods}

\begin{table}[t]
\centering
\small
\begin{tabular}{l|cccc}
    \hline
    & \multicolumn{2}{c}{\textbf{Missing Boundaries}} 
    & \multicolumn{2}{c}{\textbf{Gaussian Blur}} \\
    \cline{2-5}
    & $\delta e_x^*$ & $\delta e_v^*$ & $\delta e_x^*$ & $\delta e_v^*$ \\
    \hline
    Pure Pursuit   & $0.4492$ & $0.0265^-$ & $0.3170$ & $0.0920^-$ \\
    PD - Error     & -- & $0.0499^-$ & -- & $0.0188^-$ \\
    $\pi_{ICG}$   & -- & $0.007^-$ & $0.01^-$ & -- \\
    $\pi_{W_pG}$   & $0.4452$ & $0.1197^-$ & $1.025$ & $0.011^-$ \\
    \hline
\end{tabular}
\caption{Comparison of error deviations, $\delta e_x$ and $\delta e_v$ (difference between the tracking errors between the baseline and evaluation scenario) for real-time challenges such as missing lane boundaries (MB) and Gaussian Blur (GB)}
\label{tab:classical_compare_real}
\end{table}

Similar to simulation based evaluations, the proposed method, $\pi_{W_pG}$, is compared with existing formal navigation methods, but with a specific focus on sensitivity to variations in the input image features. For this evaluations, two scenarios as illustrated in the fig.~\ref{fig:eval_scenarios} are created where the baseline track is, (a) modified by removing cones from sections of the track to create missing lane boundaries (MB), and, (b) introducing a Gaussian blur (GB) on the input image to emphasize out of focus vision sensor or lack of distinct visual markers. Similar to simulation evaluations outlined in sec.~\ref{subsec:sim_compare_formal}, N-MPC, Pure pursuit, PD-Center tracking and $\pi_{ICG}$ where originally selected for the investigation. Unfortunately, owing to execution delays on the hardware platform, N-MPC didn't perform sufficiently well to be included in the study. 

Fig.~\ref{fig:eval_scenarios} (bottom row) indicate the trajectories realized by the controllers when the input feature gets affected in comparison to their original performance. The crosses indicate that the robot either leaving the track, or the requiring re-alignment on the track. Among all methods, only Pure pursuit provides tracking performance comparable to our proposed method ($\pi_{W_pG}$) as shown in the fig.~\ref{fig:eval_scenarios} and table~\ref{tab:classical_compare}. Unfortunately, it fails intermediately due to missing input features and the robot needs resets when faced with non-ideal vision data. $\pi_{W_pG}$ on the other hand slightly lacks in the center line tracking performance, but is robust to non-ideal input features due to its training on low dimension latent images and partial curves with lane center implicitly encoded via pose information. 

Table~\ref{tab:classical_compare_real} provides the metrics on tracking error deviations ($\delta e^*_x$ and $\delta e^*_v$) which is indicative of how much the performance worsens for every controller as compared to the ideal case. Controllers which rely purely on manipulation of pixel data such as $\pi_{ICG}$ and PD-Center tracking perform much worse with often going out of track bounds. A common challenge witnessed with PD-Center tracking is that when one of the lane goes missing, the the centroid shifts to the visible lane and the robot ends up traversing the single lane. The missing entries from the table are indicative that the run wasn't long enough to calculate a statistically significant error metric.

\section{Generalization beyond cones}~\label{sec:generalization}

\begin{figure*}
    \centering
    \includegraphics[width=0.99\linewidth]{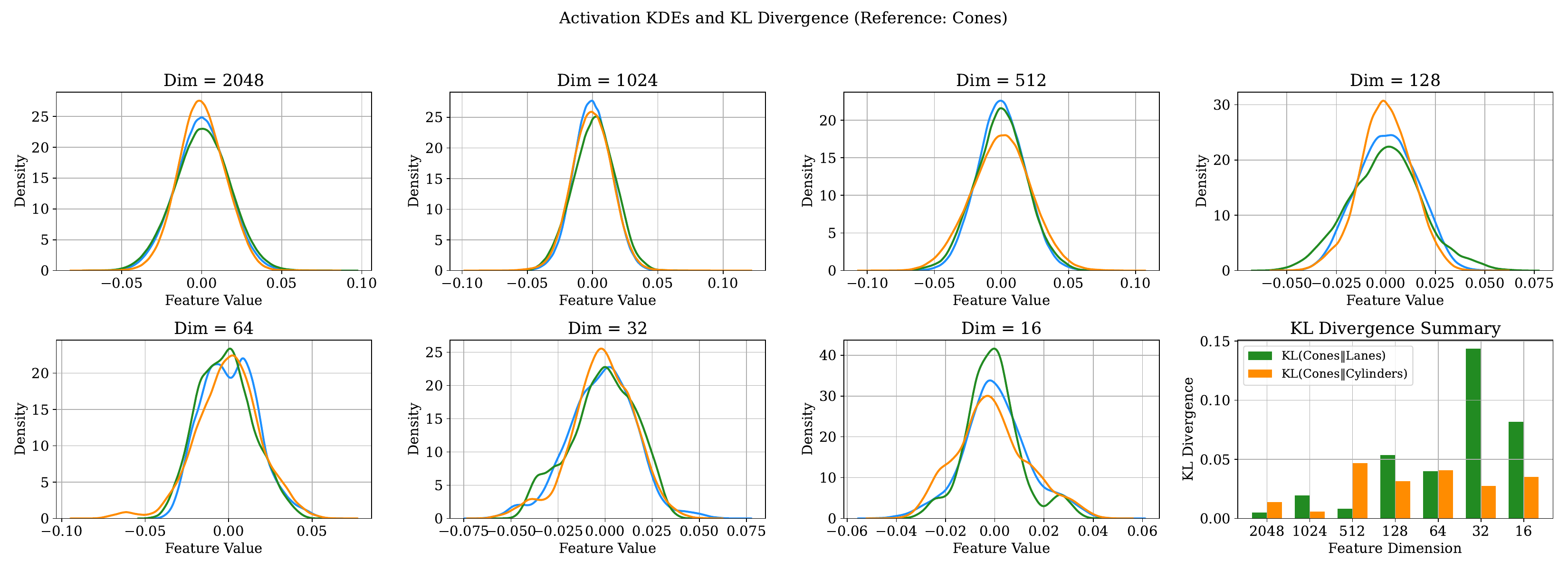}
    \caption{Feature value distributions for feature size varying from 2048 to 16 for visual markers cones, cylinders and solid lanes. KL divergence plot (bottom right) indicated feature size 64 as the the minimum number of feature dimension where both types of features have low KL divergence as compared to cones.}
    \label{fig:input-featue-distribution}
\end{figure*}

\begin{table*}
\centering
\renewcommand{\arraystretch}{1.3}
\begin{tabular}{l|cc|cc|cc|cc|cc}
\hline
\textbf{Dataset} & 
\multicolumn{2}{c|}{$\kappa$ [0.00108, 0.20086]} & 
\multicolumn{2}{c|}{$\kappa$ [0.20086, 0.40065]} & 
\multicolumn{2}{c|}{$\kappa$ [0.40065, 0.60043]} & 
\multicolumn{2}{c|}{$\kappa$ [0.60043, 0.80022]} & 
\multicolumn{2}{c}{$\kappa$ [0.80022, 1]} \\
& $e_x$~(m) & ${}^{B}V$~(m/s) & $e_x$~(m) & ${}^{B}V$~(m/s) & $e_x$~(m) & ${}^{B}V$~(m/s) & $e_x$~(m) & ${}^{B}V$~(m/s) & $e_x$~(m) & ${}^{B}V$~(m/s) \\
\hline
cones     & 0.073747 & 0.75529 & 0.062732 & 0.76961 & 0.080191 & 0.76729 & 0.12519 & 0.76622 & 0.072137 & 0.76966 \\
cylinders & 0.1624   & 0.76282 & 0.13271  & 0.7656  & 0.095616 & 0.76666 & 0.15769 & 0.73118 & 0.26513  & 0.77091 \\
lanes     & \textbf{0.055348} & 0.77424 & \textbf{0.04389}  & 0.77926 & \textbf{0.051555} & 0.7903  & \textbf{0.087934} & 0.80685 & \textbf{0.060343} & 0.78772 \\
\hline
\end{tabular}
\caption{Mean tracking error ($e_x$) and mean linear velocity (${}^{B}V$[desired to be $0.75$m/s]) for different markers (cone, cylinders and solid lanes) binned as per track curvature ($\kappa~[1/m]$).}
\label{tab:out_dist_sim_kappa}
\end{table*}

In this final section, the proposed policy $\pi_{W_pG}$ is evaluated for performance beyond the training visual markers (cones). The intuition behind this investigation is that the Atari CNN based neural network utilized in this framework for reducing the image feature dimensions makes the specific shape of the cones irrelevant an extracts only the relevant features associated with path curvature. 

Figure~\ref{fig:out-of-training} illustrates the various scenarios in simulation and reality where the policy is evaluated in a zero-shot manner. The input images undergo a simple Canny edge detection formulation to identify apparent lane boundary or some partial lane-line structure in the environment~\cite{canny1986computational}. With that as an input, the policy is able to follow the complete or partial lanes without any challenges indicating strong generalization of the policy. The two critical aspects for this generalization are (a) low dimension latent features of the input image, that allows the policy to work on any shape lane markers, and, (b) the pose information introduced in training, that allows the policy to function on single lanes or partial lane boundaries as the the visual marker-robot position relationship is encoded in the policy.

The formal intuition as to why the policy can perform on out-of-domain data is due to the fact that the visual features are distilled just enough that the cone shapes is irrelevant but the curvature specific information is retained. To test this hypothesis, over $36000$ synthetic images captured from the track were distilled in different size features using the feature extraction network. The distilled features coming from cones, cylinders and solid lanes where then analyzed to see how the control policy perceives them.

Figure~\ref{fig:input-featue-distribution} illustrates this by showing how two different visual features (cylinders and cones, as shown in fig.~\ref{fig:out-of-training}) result in different distributions of the input activation. It can be observed that for certain feature sizes, the different visual markers result in different distributions indicating that the policy would be sensitive to training features. From the KL-Divergence calculations (fig~\ref{fig:out-of-training}, bottom right) between the distribution achieved by cones and those by lanes and cylinders as markers, it is clear that feature size $64$ is the one that provides the minimum feature size vs low KL-Divergence tradeoff.  

Table~\ref{tab:out_dist_sim_kappa} captures the policy performance on tracks laid down by different visual markers. The categorization is done by binning sections of track in increasing order of the track curvature ($\kappa~[1/m]$). While in general it is noticeable that the policy performance decreases as the track curvature increases, there is no noticeable performance difference due to the usage of different markers. On the contrary, it can be seen that for all curvatures, solid lanes as visual markers out perform the usage of cones for position tracking performance, $e_x$.
\section{Discussion}~\label{sec:discussion}

A framework for pose based visual navigation for skid-steered lane keeping has been proposed in this work. The framework is extensively validated in simulation and reality for investigating its performance due to algorithmic modifications and deployment uncertainties such as sensor malfunctions. The investigation also compares the proposed framework with the widely popular approaches such as N-MPC, Pure pursuit, Proportional- Derivative based center tracking, and, image centroid guided learning for lane following.
While there is room for engineering several of the above approaches to improve their relative performance, they still demand utilizing learning based frameworks to deal with challenges of lack of environment structure, for which the proposed method triumphs.

The contribution of this work, more than the metricized performance improvements is in providing a scientifically sound methodology for formalizing the learning for lane keeping tasks, which was an existing research gap. While this framework is complete in itself, certain assumptions such as traversal on flat terrains and avoidance of zero-radius turns in can be an interesting topic of investigation for further testing the limits of this framework. 

The final section of the work indicates how training on the latent feature space allows for the policy to be marker agnostic and be deployed in various real use cases. With that in mind, integrating the policy as a module within an autonomy stack would be a good future research direction to investigate how the policy performs at a system's level. At system's level, the policy can further be improved for varying system's level behaviors such as changing robot mass, camera orientation, or traction parameters. Learning family of policies using contemporary tools such as diffusion models is another potential avenue that can be explored. 


\section*{ACKNOWLEDGMENT}

This work was supported by the Virtual Prototyping of Autonomy Enabled Ground Systems (VIPR-GS), a US Army Center of Excellence for modeling and simulation of ground vehicles, under Cooperative Agreement W56HZV-21-2-0001 with the US Army DEVCOM Ground Vehicle Systems Center (GVSC).
\section*{Appendix: Benchmark Controllers}~\label{sec:Appendix}

\begin{enumerate}
    \item \textbf{PD – Error Tracking}:

    \begin{figure}[t]
        \centering
        \includegraphics[width=0.5\linewidth]{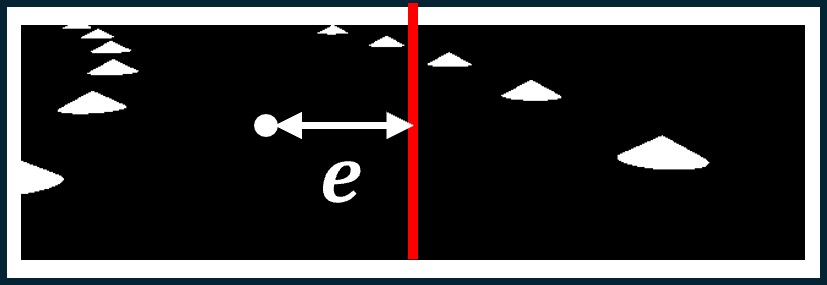}
        \caption{Lateral error ($e$) between image centroid and image frame center}
        \label{fig:centroid-error}
    \end{figure}
    
    Proportional - Derivative based error tracking is a classical apporach to track the image center such that the robot aligns itself with the lane center. An error $e$ is calculated based on the image's apparent center in the horizontal plane and is utilized as a feedback signal to generate linear and angular velocity commands as :
    \begin{equation*}
        V = V_{\text{ref}}, \quad \omega = K_p \cdot e + K_d \cdot \dot{e}
    \end{equation*}
    where, $K_p$ and $K_d$ are the proportional and derivative gains for the controller which are typically tuned for different reference velocities.

    \item \textbf{Image centroid based RL ($\pi_{ICG}$)}:

    Similar to PD based centroid tracking, the policy tries to minimized the image centering error. On contrary to gains that are tuned heuristically, $\pi_{ICG}$ tries to learn a policy based on the following reward:
    
    \begin{align*}
    \mathbf{R}_{ICG} = &\ (1 - e_{c})^{2} + (1 - e_{V})^{2} \\
                       &+ (1 - e_{a_{1}})^{2} + (1 - e_{a_{2}})^{2}
    \end{align*}

    where \(e_{c} \in [0,1]\) represents the normalized lateral lane centering error, which is the deviation of the image centroid from the vertical central axis of the image plane.

    \item \textbf{Nonlinear Model Predictive Control (N-MPC)}:
    \begin{figure}
        \centering
        \includegraphics[width=0.95\linewidth]{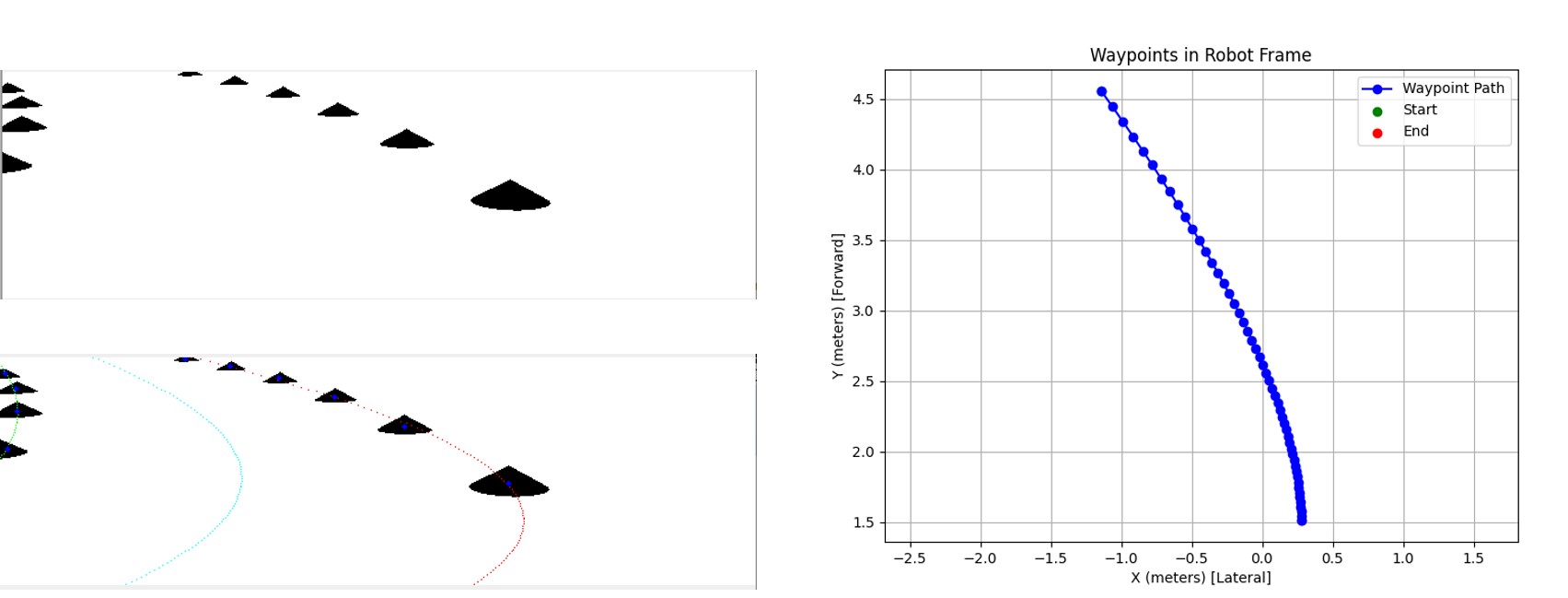}
        \caption{(Left)Second order quadratic polynomial fit via lane boundaries to find lane center points identified as $U$. (Right) Lane center points converted to waypoints ($X$) in body frame via homograph transformation ($H$) as $X = HU$.}
        \label{fig:enter-label}
    \end{figure}
    
    Non-linear model predictive controller adopts the nominal robot model to identify optimum control actions for tracking a reference trajectory~\cite{LK-MPC}. The non-linear differential drive model represented by :

    \begin{align*}
        \begin{bmatrix}
            \dot{X} \\
            \dot{Y} \\
            \dot{\theta}
        \end{bmatrix} =
        \begin{bmatrix}
            \cos\theta & -\sin\theta &0\\
            \sin \theta & \cos\theta &0 \\
            0 & 0& 1
        \end{bmatrix}
    \end{align*}

    is utilized at by linearization at every timestep. The descrite time model thus obtained $f_d$ is leveraged to formulate the controller to evaluate $N$ control actions $\mathbf{u} : [V,\omega]$: 
    
    \begin{align*}
        &\min_{\mathbf{u}_{0:N}} \sum_{k=0}^{N} \left\| \mathbf{x}_k - \mathbf{x}_{\text{ref},k} \right\|_Q^2 + \left\| \mathbf{u}_k \right\|_R^2 \\ \quad
        &\text{s.t.} \quad \mathbf{x}_{k+1} = f_d(\mathbf{x}_k, \mathbf{u}_k) \\
        &\text{subject to} \quad V \in [0,1] \quad \text{and} \quad \omega \in [-0.5,0.5]
    \end{align*}

    \item \textbf{Pure Pursuit}:
    Pure pursuit is a geometric path following controller that generates the angular velocity commands such that the robot traces a smooth arc to the waypoint of choice $\mathbf{X}$~\cite{LK-PurePursuit}. The waypoint of choice, $\mathbf{X}$ is chosen using a set look-ahead distance, $L$ which is tuned heuristically based on robot's velocity. The robot's linear velocity, $V$ is kept fixed and the angular velocity, $\omega$ is calculated as : 
    
    \begin{align*}
        &\omega = \frac{2 V \sin(\alpha)}{L} \\
        & \text{where}, \alpha = \tan^{-1}\left(\frac{y}{x}\right), \quad     
    \end{align*}
    $x$ and $y$ are the coordinates of the tracking waypoint $\mathbf{X}$.
\end{enumerate}

\bibliography{references,bib_skid_steer}

\end{document}